\documentclass[10pt,twocolumn,letterpaper]{article}

\usepackage[pagenumbers]{iccv} %
\usepackage{graphicx}
\usepackage{placeins}
\usepackage{multirow}
\usepackage{comment}

\usepackage[normalem]{ulem}
\useunder{\uline}{\ul}{}
\usepackage{mathtools}
\usepackage{tikz}
\def\checkmark{\tikz\fill[scale=0.4](0,.35) -- (.25,0) -- (1,.7) -- (.25,.15) -- cycle;} 
\usepackage{subcaption} %

\newcommand{\rebbAdd}[1]{#1}

\usepackage{amsmath,amsfonts,bm}

\def\eqref#1{equation~\ref{#1}}
\def\Eqref#1{Equation~\ref{#1}}

\def\1{\bm{1}}

\DeclareMathAlphabet{\mathsfit}{\encodingdefault}{\sfdefault}{m}{sl}
\SetMathAlphabet{\mathsfit}{bold}{\encodingdefault}{\sfdefault}{bx}{n}

\definecolor{iccvblue}{rgb}{0.21,0.49,0.74}
\usepackage[pagebackref,breaklinks,colorlinks,allcolors=iccvblue]{hyperref}

\def\paperID{2626} %
\def\confName{ICCV}
\def\confYear{2025}

\title{Utilization of Neighbor Information for Image Classification with Different Levels of Supervision}
\author{Gihan Jayatilaka \quad Abhinav Shrivastava \quad Matthew Gwilliam\\
University of Maryland\\
{\tt\small \{gihan, abhinav, mgwillia\} @cs.umd.edu}
}

\begin{document}
\maketitle

\begin{abstract}
    We propose to bridge the gap between semi-supervised and unsupervised image recognition with a flexible method that performs well for both generalized category discovery (GCD) and image clustering.
    Despite the overlap in motivation between these tasks, the methods themselves are restricted to a single task -- GCD methods are reliant on the labeled portion of the data, and deep image clustering methods have no built-in way to leverage the labels efficiently.
    We connect the two regimes with an innovative approach that \textbf{U}tilizes \textbf{N}eighbor \textbf{I}nformation for \textbf{C}lassification (\textbf{UNIC}) both in the unsupervised (clustering) and semisupervised (GCD) setting.
    State-of-the-art clustering methods already rely heavily on nearest neighbors.
    We improve on their results substantially in two parts, first with a sampling and cleaning strategy where we identify accurate positive and negative neighbors, and secondly by finetuning the backbone with clustering losses computed by sampling both types of neighbors.
    We then adapt this pipeline to GCD by utilizing the labelled images as ground truth neighbors.
    Our method yields state-of-the-art results for both clustering (+3\% ImageNet-100, Imagenet-200) and GCD (+0.8\% ImageNet-100, +5\% CUB, +2\% SCars, +4\% Aircraft).
  
\end{abstract}

\addtocontents{toc}{\protect\setcounter{tocdepth}{-1}}
\section{Introduction}

\begin{figure*}
    \centering
    \includegraphics[width=0.8\textwidth]{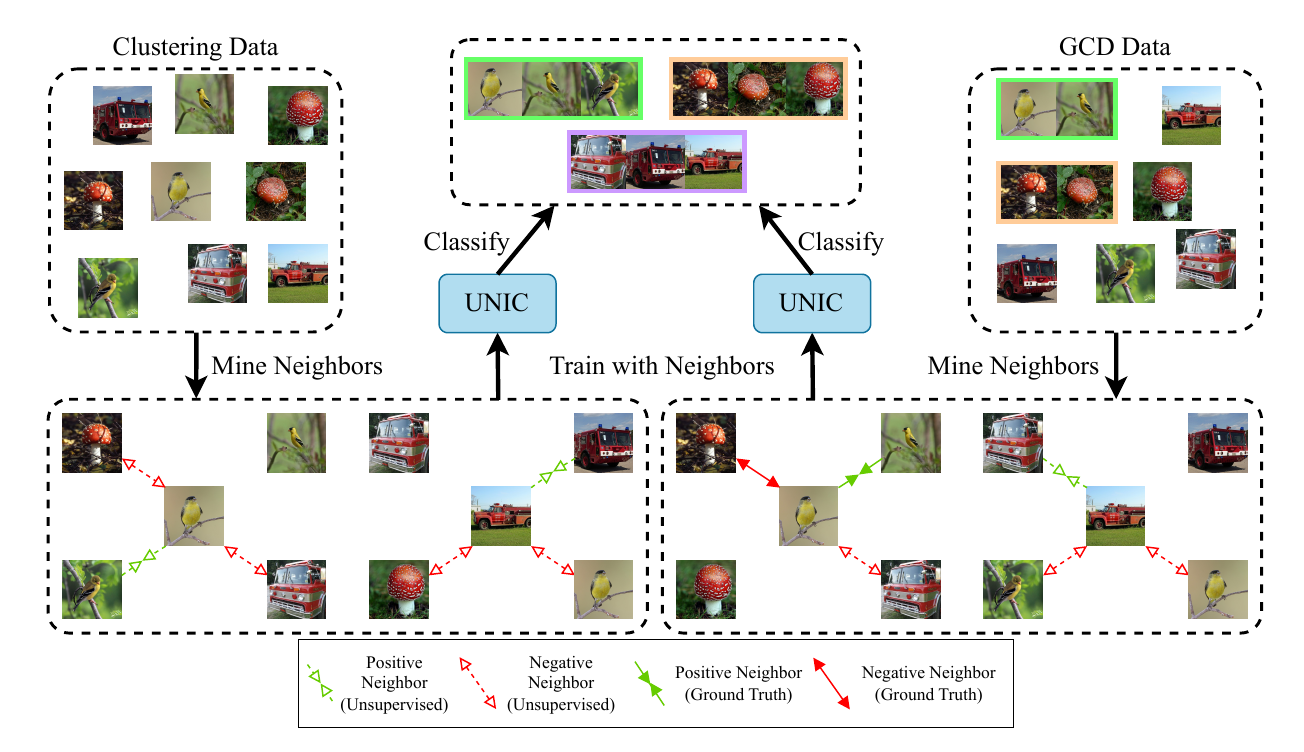}
    \caption{\textbf{Unifying Clustering and GCD.} We observe that the goals of image clustering and generalized category discovery (GCD) are identical, they only differ slightly in terms of supervision (top). Therefore, we propose a clustering approach based on mining of positive and negative neighbors, which belong to the same class as an anchor and a different class, respectively (bottom left). We can extend this approach for GCD by using the ground truth labels for ``perfect'' neighbors (bottom right).}
    \label{fig:teaser}
\end{figure*}

Image recognition has long been treated as a fundamental task in computer vision. 
One very popular setting is supervised image classification, where each image has a single ground truth label, and a model learns to predict labels by training with image-label pairs~\citep{5206848}.
As work in the fully-supervised setting plateaus, interest in the field has shifted to settings more reminiscent of the real world, where access to labeled data is more limited, or even completely absent.
So, many different recognition tasks have evolved based on the presence or absence, fraction of presence, quality, granularity, and type of labels~\citep{koch2015siamese,Bendale_2015,snell2017prototypical,vinyals2017matching,suri2023sparsedet}. 
At the extreme, unsupervised classification~\citep{SCAN_ECCV2020}, or deep image clustering, deals with the case where none of the labels are disclosed during the training step. 
Recently, generalized category discovery (GCD)~\citep{GCD_CVPR2022} has been proposed, such that some images from some classes are labelled, and some classes are completely unlabelled. 

We propose a method, UNIC (pronounce ``you-neek''), which can solve both clustering and GCD, as shown in Figure~\ref{fig:teaser}.
When labels are unavailable, UNIC can operate as a state-of-the-art deep image clustering pipeline.
Where labels are available, UNIC can leverage them for SOTA performance on GCD.
We accomplish this by designing UNIC to exploit neighbor information in the feature space of high quality unsupervised image representation models.
For every image, we mine both ``positive'' (belongs to the same class) and ``negative'' (belongs to a different class) neighbors.
We then finetune the unsupervised model end-to-end with losses that pull the positive neighbors together, and push the negative neighbors apart.
This adapts quite naturally to GCD, since we can use the labels to guarantee 100\% accuracy for a portion of the positive and negative neighbors.

We find that with the unsupervised DINO~\citep{caron2021emerging} ViT~\citep{dosovitskiy2021image} backbone, we can reliably extract both positive and negative neighbors.
The success of our method is fully dependent on the quality of these neighbors.
So, we carefully analyze the representation space and propose a novel nearest-neighbor cleaning strategy for ``positive'' neighbors. Intuitively, images at the decision boundary will have noisy neighbors (because the neighbors belong to different classes) while images near the centroid of classes will have cleaner neighbors. We identify the union size of second-order neighbors (the neighbors of neighbors) as a proxy for this purity measure and discard others.

We improve on prior clustering work that relies on a three-stage pipeline~\citep{SCAN_ECCV2020} with our two-stage pipeline, where aside from the initial representation learning, our clustering is learned in one step.
After neighbor mining, we finetune the model end-to-end.
Class assignment is learned with a clustering loss.
Meanwhile, we use a maximum entropy loss to ensure that the clustering does not collapse to a trivial solution.
Due to the strength of the DINO backbone itself, the influence of the negative neighbors, and the fact that we are able to successfully finetune, we find that the self-labeling step is not necessary, and gives diminishing returns.

Once we establish the good clustering properties of the method, we apply it to the GCD task.
GCD methods typically rely on some clustering for initialization~\citep{GCD_CVPR2022,DCCL_CVPR2023}.
However, we find such initialization totally unnecessary as we can instead learn a partially-supervised clustering end-to-end.
That is, we are able to treat the labelled images as a special case where we simply supervise their neighbor mining.

Strikingly, this results in much better performance on the labeled classes for other method compared to others, while also having competitive results in unlabeled classes.

We summarize our contributions as follows:

\begin{itemize}
    \item We propose a novel neighbor mining strategy where we improve the accuracy of positive neighbors via cleaning and introduce the idea of negative neighbors for image clustering.
    \item We formulate a general pipeline which can be trained end-to-end for image clustering and, with minimal adaptation, for the semi-supervised GCD task as well.
    \item We achieve SOTA on both tasks, unifying classification for unsupervised and semi-supervised methods, with consistent gains over most datasets in both GCD and Clustering.
\end{itemize}

\begin{figure*}[htb]
    \begin{centering}
    \includegraphics[width=0.8\textwidth]{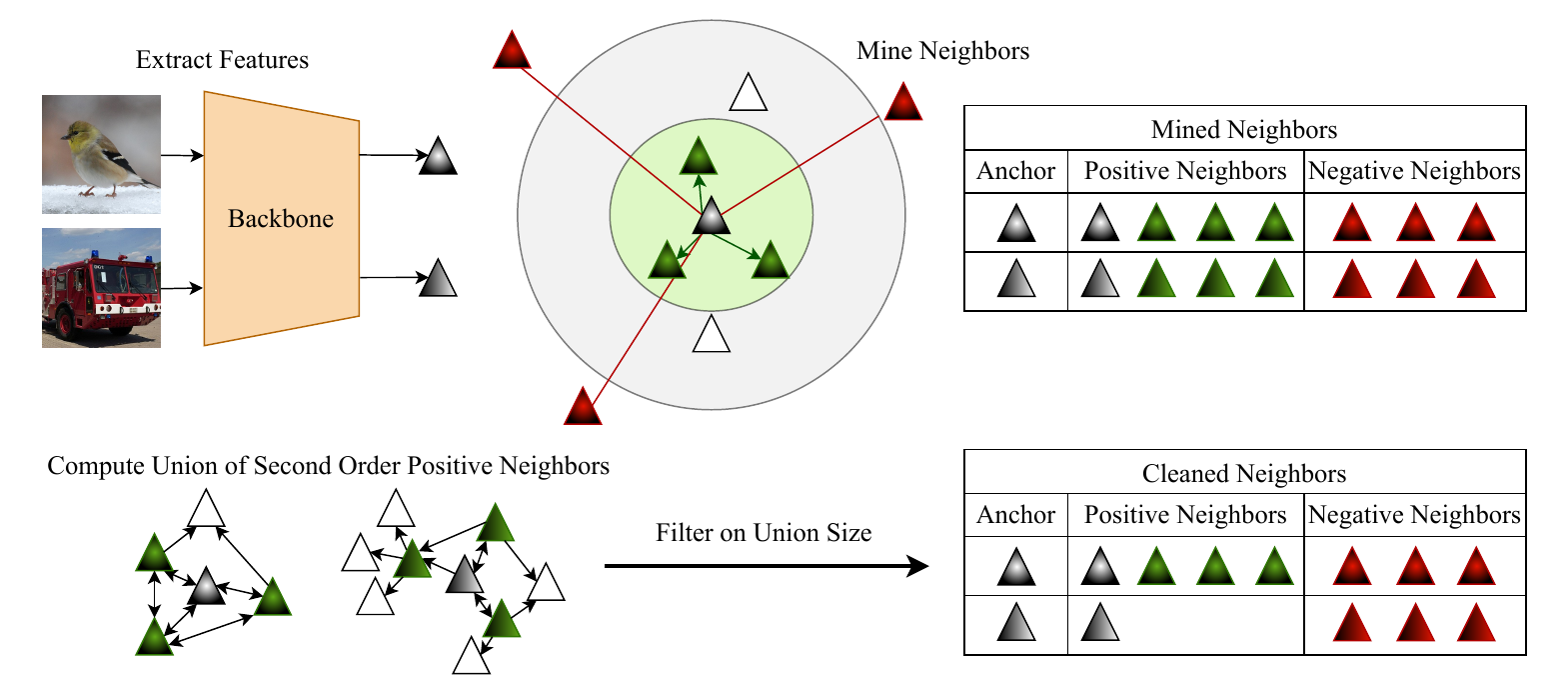}
    \caption{\textbf{Neighbor Mining} for UNIC. We extract features from a backbone, take the closest samples as ``positive'' neighbors, and some of the far samples for ``negative'' neighbors. We then prune some positive neighbors, depending on the number of mutual nearest neighbors (union of nearest neighbors of nearest neighbors). \label{fig:mining-diagram}}
    
    \end{centering}
\end{figure*}
\begin{figure*}[htb]
    \centering
    \includegraphics[width=0.8\textwidth]{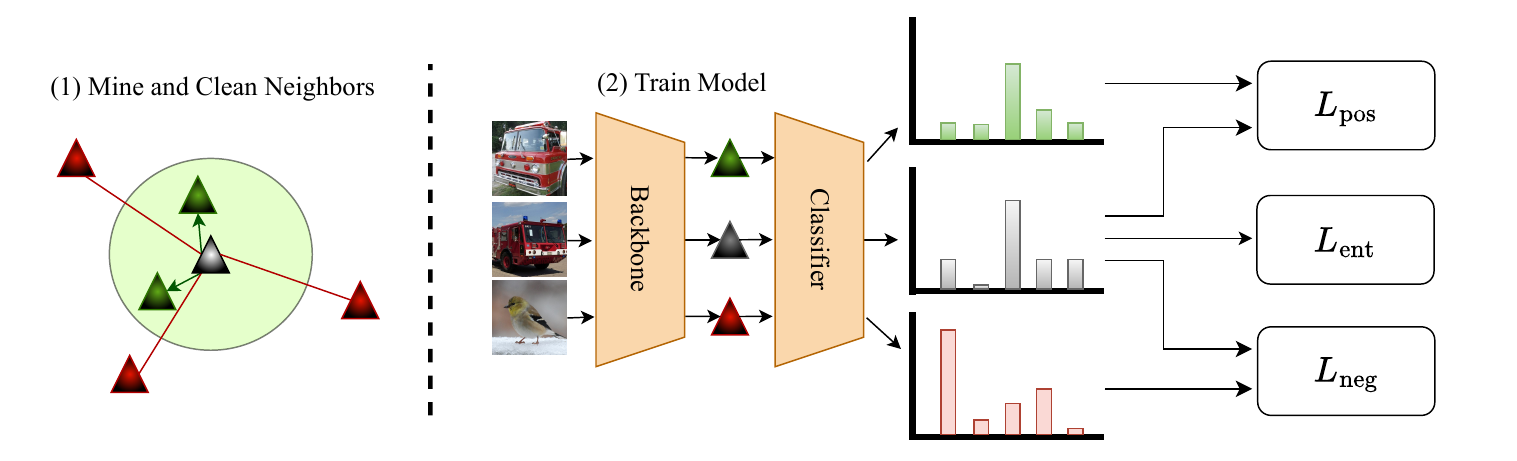}
    \caption{\textbf{UNIC}. We first mine neighbors (see Figure~\ref{fig:mining-diagram}). We finetune the backbone with a classification head that we train without labels, using losses that encourage the model to predict the same class for positive neighbors, different classes for negative neighbors, and entropy for regularization.}
    \label{fig:method-diagram}
\end{figure*}

\section{Related Work}

\subsection{Unsupervised Representation Learning} 
\label{sec:Unsupervised-Representation-Learning} 

Unsupervised representation learning refers to the process of training deep neural networks to extract information from images without labels.
Some early approaches rely on a host of diverse pretext tasks -- patch prediction~\citep{doersch2015unsupervised,gupta2020patchvae}, jigsaw solving~\citep{noroozi2016unsupervised}, colorization~\citep{larsson2016learning,zhang2016colorful}, inpainting~\citep{pathak2016context}, rotation prediction~\citep{gidaris2018unsupervised}, image generation~\citep{donahue2019large}, etc.~\citep{doersch2017multi}.
Contrastive~\citep{Wu_2018_CVPR,pmlr-v119-chen20j,chen2020improved,misra2020self,bachman2019learning,chen2020simple,DBLP:journals/corr/abs-1911-05722,hjelm2018learning,henaff2020data,oord2018representation,tian2020contrastive,Ye_2019_CVPR,DBLP:journals/corr/abs-2103-03230,chen2021exploring} and clustering~\citep{caron2018deep,asano2019self,caron2019unsupervised,caron2020unsupervised,caron2021emerging,li2021prototypical} methods have largely superseded these.
More recent approaches rely on image reconstruction~\citep{DBLP:journals/corr/abs-2111-06377,assran2022masked,zhou2022ibot,huang2022contrastive,bao2022beit,mishra2022simple} and generation~\citep{mukhopadhyay2023textfree,hudson2023soda,li2022mage}.
Currently these clustering, contrastive, and reconstruction-based approaches are all among the state-of-the-art.
In our work, since we do deep clustering (an unsupervised problem), we need backbones which do not require labels, and we primarily use a ViT-B~\citep{dosovitskiy2021image} trained with DINO~\citep{caron2021emerging} due to the nice properties of its embeddings with respect to nearest neighbors. Hard negative mining ~\citep{mochi,robinson2021contrastive} has shown improvements ~\citep{Wang2020UnderstandingCR} in learned representations for fully supervised downstream tasks. 

\subsection{Deep Clustering}

Deep clustering aims to learn groups of images from unlabelled data, where the images in a given grouping (cluster) have similar visual semantics. 
A naive approach would use kmeans on top of some embeddings extracted by a model trained with unsupervised representation learning, and indeed, this task has been used as a benchmark for the quality of such representations~\citep{gwilliam2022supervised}.
The work in this area can be primarily divided between single-stage and multi-stage methods, as in \citep{TEMI_BMVC2023}.
Some of these target the clustering problem directly~\citep{Chang_2017_ICCV}, but many of the single-stage methods use clustering primarily as a mechanism for learning good image representations~\citep{caron2019unsupervised,caron2020unsupervised,asano2019self}.
Indeed, these methods are only single-stage in the sense they have one learning stage, and they often still rely on running k-means on the final representations for the actual cluster predictions~\citep{li2021prototypical,ProPos_IEEETrans2023}.

The multi-stage pipeline is first proposed with SCAN~\citep{SCAN_ECCV2020}, which starts with a representation learning stage, followed by a stage where a clustering head is learned based on nearest neighbor mining, then a final stage where the head is further tuned with pseudo ground truths.
Other methods offer deviations within this pipeline -- NNM matches neighbors at both the batch and global level~\citep{Dang_2021_CVPR}.
SPICE follows similar stages, but with a heavier emphasis on pseudo-labeling~\citep{Niu_2022}.
TSP~\citep{zhou2022deep} and TEMI~\citep{TEMI_BMVC2023} leverage ViTs to outperform the earlier methods (which mainly use ResNets~\citep{he2015deep}), although it is worth noting that all these methods are conceptually backbone-agnostic.
Our method is conceptually aligned with SCAN and NNM except that we introduce key novelties with respect to the neighbor mining, namely with our process for cleaning positive neighbors, and our use of negative neighbors.
Also, compared to many of these methods, we use a single clustering head for less expensive training.

\subsection{Generalized Category Discovery} 

GCD is an open world image recognition task~\citep{GCD_CVPR2022}.
To set up some GCD task, we typically take an existing image dataset and cut the classes in half, with some known, the other unknown.
We then use labels for half of the images for the known classes.
The same problem is often sometimes referred to as open-world semi-supervised learning~\citep{cao2021open}. 
The preliminary solution (also referred to as GCD) combines unsupervised contrastive learning for the unlabelled images and supervised contrastive learning for the labelled images, along with an extension of k-means to support the labelled classes~\citep{GCD_CVPR2022}.
DCCL~\citep{DCCL_CVPR2023} uses a two alternating stages in which visual concepts are discovered and then images are represented with these concepts. 
SimGCD~\citep{Wen_2023_ICCV} mitigates prediction biases in parametric classifiers with entropy regularization.
PromptCAL~\citep{Zhang_2023_CVPR} utilizes graph structures for a supervisory signal. 
GCP~\citep{GCP_ICCV2023} uses Gaussian Mixture Models.
Our method, UNIC is a parametric approach with a learnable clustering head where we reinforce connections between neighbors to learn both labelled and unlablled classes and images.

\begin{table*}[t]
\centering
\begin{minipage}{0.75\linewidth}
\resizebox{\textwidth}{!}{%
\begin{tabular}{@{}ll ccc ccc ccc ccc@{}}
\toprule
 & & \multicolumn{3}{c}{STL-10} & \multicolumn{3}{c}{ImageNet-50} & \multicolumn{3}{c}{ImageNet-100} & \multicolumn{3}{c}{ImageNet-200} \\
 \cmidrule(l){3-5}
 \cmidrule(l){6-8}
 \cmidrule(l){9-11}
 \cmidrule(l){12-14}
Algorithm & Backbone & ACC & NMI & ARI & ACC & NMI & ARI & ACC & NMI & ARI & ACC & NMI & ARI \\
\midrule
kMeans~\cite{SCAN_ECCV2020} & ResNet & 65.8 & 60.4 & 50.6 & 65.9 & 77.5 & 57.9 & 59.7 & 76.1 & 50.8 & 52.5 & 75.5 & 43.2 \\MoCoV2\textsuperscript{\textdagger}~\cite{chen2020improved} & \rebbAdd{ResNet} & \rebbAdd{71.81} & \rebbAdd{66.52} & \rebbAdd{52.54}  & \rebbAdd{63.04} & \rebbAdd{75.75} & \rebbAdd{47.00} & \rebbAdd{60.30} & \rebbAdd{75.13} & \rebbAdd{42.53} & \rebbAdd{52.35} & \rebbAdd{73.52} & \rebbAdd{37.21} \\
SCAN~\cite{SCAN_ECCV2020} & ResNet & 80.9 & 69.8 & 64.6 & 76.8 & 82.2 & 66.1 & 68.9 & 80.8 & 57.6 & 58.1 & 77.2 & 47.0 \\
ProPos~\cite{ProPos_IEEETrans2023} & ResNet-50 & 86.7 & 75.8 & 73.7 & - & 82.8 & 69.1 & - & 83.5 & 63.5 & - & 80.6 & 53.8 \\
DPM~\cite{deepDPM_CVPR2022} & ResNet & 85.0 & 79.0 & 71.0 & 66.0 & 77.0 & 54.0 & - & - & - & - & - & - \\
DAC~\cite{dac_ICCV2017} & ResNet-50 & 47.0 & 36.6 & 25.7 & - & - & - & - & - & - & - & - & - \\
\rebbAdd{HCL\textsuperscript{\textdagger}}~\cite{robinson2021contrastive} & \rebbAdd{ResNet} & \rebbAdd{62.34} & \rebbAdd{60.17} & \rebbAdd{43.42}  & - & - & - & - & - & - & - & - & - \\
\rebbAdd{MoCHi\textsuperscript{\textdagger}}~\cite{mochi} & \rebbAdd{ResNet} & \rebbAdd{69.79} & \rebbAdd{64.00} & \rebbAdd{50.32}  & \rebbAdd{61.88} & \rebbAdd{73.44} & \rebbAdd{44.92} & \rebbAdd{57.72} & \rebbAdd{73.44} & \rebbAdd{42.48} & \rebbAdd{49.16} & \rebbAdd{70.79} & \rebbAdd{33.38} \\
\midrule
TSP~\cite{zhou2022deep} & ViT-B/16 & 97.9 & 95.8& 95.6& -& -& -& -& -& -& -&- & -\\
TEMI~\cite{TEMI_BMVC2023} & ViT-B/16 & 98.5 & 96.5 & 96.8 & 80.01 & 86.10 & 70.93 & 75.05 & 85.65 & 65.45 & 73.12 & 85.20 & 62.13 \\
kMeans\textsuperscript{\textdagger} & ViT-B/16 & 97.10 & 94.02& 93.62 & 82.36 & 87.91 & 73.89 & 76.88 & {\ul 86.93} & 68.01 & 70.58 & 84.58 & 59.94 \\
TEMI\textsuperscript{\textdagger}  & ViT-B/16(Tune) & {\ul 98.65} & {\ul 96.70} & {\ul 97.04} & 80.12 & 86.30 & 70.64 & 76.80 & 86.41 & 67.42 & {\ul 73.69} & {\ul 85.72} & {\ul 63.16} \\
SCAN\textsuperscript{\textdagger} & Vit-B/16 & - & - & - & {\ul 85.48} & {\ul 88.58} & {\ul 78.19} & {\ul 77.84} & 85.97 & {\ul 68.75} & 72.89 & 84.20 & 61.85 \\
\midrule
UNIC (Ours) & ViT-B/16 & \textbf{98.75} & \textbf{96.87} & \textbf{97.25} & \textbf{90.80} & \textbf{91.81} & \textbf{84.25} & \textbf{80.84} & \textbf{88.13} & \textbf{72.70} & \textbf{75.25} & \textbf{85.93} & \textbf{64.73} \\

\bottomrule
\end{tabular}
}
\end{minipage}
\caption{\textbf{Image Clustering Results.} We achieve SOTA on accuracy, NMI, and ARI for all datasets. \textsuperscript{\textdagger} denotes our implementations. \label{tab:results_clustering}
}

\end{table*}

\begin{table*}[t]
\centering
\begin{minipage}{0.75\linewidth}
\resizebox{\textwidth}{!}{%
\begin{tabular}{@{}ll ccc ccc ccc ccc ccc@{}}
\toprule
& & \multicolumn{3}{c}{ImageNet-100} & \multicolumn{3}{c}{CUB-200} & \multicolumn{3}{c}{CIFAR-10} & \multicolumn{3}{c}{Aircrafts} & \multicolumn{3}{c}{SCars} \\ 
\cmidrule(l){3-5}
\cmidrule(l){6-8}
\cmidrule(l){9-11}
\cmidrule(l){12-14}
\cmidrule(l){15-17}
Algorithm & Backbone & All & Old & New & All & Old & New & All & Old & New & All & Old & New & All & Old & New \\ \midrule
kMeans ~\cite{GCD_CVPR2022} & DINOv1 & 72.2 & 75.5 & 71.3 & 34.3 & 38.9 & 32.1 & 83.6 & 85.7 & 82.5 & 12.9 & 12.9 & 12.8 & 12.8 & 10.6 & 13.8 \\
UNO+~\cite{fini2021unified} & DINOv1 & 70.3 & 95.0 & 57.9 & 35.1 & 49.0 & 28.1 & 68.6 & 98.3 & 53.8 & 28.3 & 53.7 & 14.7 & 35.5 & 70.5 & 18.6 \\
ORCA~\cite{cao2021open} & DINOv1 & 73.5 & 92.6 & 63.9 & 36.3 & 43.8 & 32.6 & 81.8 & 86.2 & 79.6 & 31.6 & 32.0 & 31.4 & 31.9 & 42.2 & 26.9 \\
GCD~\cite{GCD_CVPR2022} & DINOv1 & 74.1 & 89.8 & 66.3 & 51.3 & 56.6 & 48.7 & 91.5 & 97.9 & 88.2 & 45.0 & 41.1 & 46.9 & 39.0 & 57.6 & 29.9 \\
DCCL~\cite{DCCL_CVPR2023} & DINOv1 & 80.5 & 90.5 & 76.2 & 63.5 & 60.8 & 64.9 & 96.3 & \underline{96.5} & 96.9 & - & - & - & 43.1 & 55.7 & 36.2 \\
Prompt CAL~\cite{Zhang_2023_CVPR} & DINOv1 & 83.1 & 92.7 & 78.3 & 62.9 & 64.4 & 62.1 & \textbf{97.9} & \textbf{96.6} & 98.5 & 52.2 & 52.2 & 52.3 & 50.2 & 70.1 & 40.6 \\
SimGCD~\cite{Wen_2023_ICCV} & DINOv1* & 83.0 & 93.1 & 77.9 & 60.3 & 65.6 & 57.7 & 97.1 & 95.1 & 98.1 & 54.2 & 59.1 & 51.8 & 53.8 & 71.9 & 45.0 \\
SPTnet~\cite{sptnet_iclr2024} & DINOv1* & 85.4 & 93.2 & 81.4 & 65.8 & 68.8 & 65.1 & 97.3 & 95.0 & \textbf{98.6} & \underline{59.3} & \underline{61.8} & \underline{58.1} & \underline{59.0} & \underline{61.8} & \textbf{58.1} \\
\midrule
UNIC (ours) & DINOv1* & 84.55 & 93.39 & 80.11 & 69.4 & \underline{83.4} & 62.4 & \underline{97.6} & 96.2 & \underline{98.3} & \textbf{63.9} & \textbf{68.1} & \textbf{61.8} & \textbf{62.6} & \textbf{90.7} & \underline{49.0} \\
\midrule
kMeans & DINOv2 & 77.64 & 84.28 & 74.31 & 69.22 & 70.91 & 68.36   & - & -& -& -& -& -& -& -& -\\
SimGCD~\cite{Wen_2023_ICCV} & DINOv2 & 88.5 & \textbf{96.2} & 84.6 & 74.9 & 78.5 & 73.1 & - & -& -& -& -& -& -& -& -\\
SPTnet~\cite{sptnet_iclr2024} & DINOv2 & \underline{90.1} & \underline{96.1} & \underline{87.1} & \underline{76.3} & 79.5 & \underline{74.6} & - & -& -& -& -& -& -& -& - \\
\midrule
UNIC (ours) & DINOv2 & \textbf{90.86} & 95.32 & \textbf{88.63} & \textbf{81.36} & \textbf{84.52} & \textbf{79.78}  & - & -& -& -& -& -& -& -& -\\
\bottomrule
\end{tabular}%
}

\end{minipage}
{\centering \caption{\textbf{Generalized Category Discovery Results} reported for all, old, and new Classes with DINOv1 backbone. DINOv2~\cite{dinov2} results are reported for a limited set where there are comparision numbers in literature. UNIC achieves SOTA for ImageNet-100, CUB-200, Aircrafts, and SCars. * denotes pretrained as per SimGCD.}\label{tab:gcd_results}}
\end{table*}

\section{Approach}
\label{sec:method}

This section starts with a formalization of the two problem settings being addressed -- image clustering and generalized category discovery.
Then, we describe our pipeline in a few stages. The first stage is neighbor mining. 
Then, we look into designing the neural network architecture and formulating the loss function for the tasks. Finally, we describe the training process along with implementation details and hyperparameters.

\subsection{Problem Settings}

Consider a training set $D_t = X \times Y$ (where $X$ is the set of images and $Y$ is the set of labels) of $N = \|D_t\|$ images and a evaluation set $D_e$. $D_t$ consists of two subsets unlabelled training set $D_U = X \times Y_U$ and labelled training set $D_L = X \times Y_L$. 
$D_L$ is a null set for image clustering. In that case, $D_u$ is as same as $D_t$. That is, there are no labeled images.

The evaluation set $D_e$ is different for clustering and GCD settings. We use the evaluation set of the image dataset of choice (which is disjoint from the training set) for the clustering setting. However, $D_e$ is $D_U$ for the GCD setting. The objective of both clustering and GCD is to come up with a function $f$ that successfully maps $X$ to $Y$ in $D_e$. Note that $f = f_{cls}\circ f_{emb}$ for the rest of this section.

\subsection{Mining and Cleaning Neighbors}

Mining neighbors refers to embedding all the images as vectors and finding the nearest and furthest examples for every vector, as illustrated in Figure~\ref{fig:mining-diagram}. We measure distances in Euclidean space. We pick the nearest $\tau_1$ examples as positive neighbors. We consider the examples that are further than the first $\tau_2$ nearest neighbors as the negative neighbors.

Consider images $x_i \in D_t$ and an embedding function $f_{emb;\theta_e}$ parameterized by $\theta_e$. 
The embedding vectors $v_i$ are generated as per ~\Eqref{eq:embedding}. 
Let $Q(x_i)$ be a permutation of $D_t$ sorted by the Euclidean distance of their embedding vectors to $v_i$ as per ~\Eqref{eq:sorting}. 
Let $0 < \tau_1 < \tau_2 < N$ be two integer parameters. 
We pick a set of positive neighbors $N(x_i)$ and a set of negative neighbors $\overline{N}(x_i)$ for $x_i$ as per ~\Eqref{eq:nei-cutoff}.

\begin{equation}
    \label{eq:embedding}
    v_i \xleftarrow{} f_{emb;\theta_e}(x_i); \forall x_i \in D_t.
\end{equation}
\begin{align}
    \label{eq:sorting}
    Q(x_i) \xleftarrow{} argsort_{x_j}\{\|f_{emb;\theta_e}(x_j) - f_{emb;\theta_e}(x_i) \|;\\
    \forall x_j \in D_t\} \notag
\end{align}
\begin{equation}
    \label{eq:nei-cutoff}
    N(x_i), \overline{N}(x_i) \xleftarrow{} Q(x_i)[0:\tau_1] , Q(x_i)[\tau_2:N] 
\end{equation}

It should be noted that every image $x_i$ would be the closest neighbor to itself. Cleaning neighbors refers to discarding the set of $N(x_i)$'s for whom the size of second-order neighborhoods goes beyond an integer parameter $\eta$ as described in ~\Eqref{eq:clearning}. Once $N(x_i)$ is discarded, the only remaining positive neighbor for $x_i$ would be itself.

\begin{equation}
\label{eq:clearning}
N_{clean}(x_i) \xleftarrow[]{}
\begin{cases}
  N_i & \left | \bigcup\limits_{x_j \in N(x_i)}^{} {N(x_j)} \right | \leq \eta  \\
  x_i        & otherwise
\end{cases}
\end{equation}

\subsection{Training}
Consider an image $x_i$, a positive neighbor $x_p \in N_{clean}(x_i)$ and a negative neighbor $x_n \in \overline{N}(x_i)$. These are denoted by the two red trucks and the bird in Figure~\ref{fig:method-diagram}. They are embedded into $f_{emb;\theta_e}(x_i), f_{emb;\theta_e}(x_p)$ and $f_{emb;\theta_e}(x_n)$. These embedding vectors are shown by grey, green, and red triangles.
Consider a classifier function $f_{cls;\theta_c}$ parameterized by $\theta_c$ that takes in an embedding vector and outputs a probability distribution over the $K$ classes. This probability distribution will be denoted by $\hat{y}_i,\hat{y}_p,\hat{y}_n \in [0,1]^{K}$. 

\begin{equation}
    \hat{y}_i, \hat{y}_p, \hat{y}_n \xleftarrow{} f_{cls;\theta_c}(f_{emb;\theta_e}(x_j)); \forall x_j \in {x_i, x_p, x_n}
\end{equation}

Let $<y_i,y_j>$ denote the dot product between two vectors $y_i$ and $y_j$. $H_b(a,b)$ denotes the binary cross entropy between two probabilities $a$ and $b$. We calculate two losses $L_{pos}$ and $L_{neg}$ as per ~\Eqref{eq:similarity-pos-neg}.

\begin{equation}
\begin{split}
    \label{eq:similarity-pos-neg}
        L_{pos}(x_i,x_p) &= H_b(<\hat{y}_i, \hat{y}_p> , 1.0)\\
        L_{neg}(x_i,x_n) &= H_b(<\hat{y}_i, \hat{y}_n> , 0.0)
\end{split}
\end{equation}

We can calculate a loss function $L_{sim}$ for the whole dataset as per Equation~\ref{eq:lsim}

\begin{align}
    \label{eq:lsim}
    L_{sim} &= \mathbb{E}_{x_i \in D_t} \left[ \mathbb{E}_{x_p \in N_{clean}(x_i)} \left[ L_{pos}(x_i,x_p) \right] \right] \\
    &\quad + \mathbb{E}_{x_n \in \overline{N}(x_i)} \left[ L_{neg}(x_i,x_n) \right] \notag
\end{align}

In addition, we calculate the entropy of the probability distribution of classes for all the examples in the dataset as per Equation~\ref{eq:entr}. 
$H( )$ is the entropy of a probability distribution here.

\begin{equation}
    \label{eq:entr}
    L_{ent} = H \biggr(\displaystyle \mathop{\mathbb{E}}_{x_i \in D_t}{\bigl[ \hat{y_i} \bigr]} \biggr)
\end{equation}

Finally, we calculate the overall loss function dataset by taking a weighted average of the two loss functions and solve the optimization problem as shown in Equation~\ref{eq:optim}.

\begin{equation}
    \label{eq:optim}
    \theta_e^* , \theta_c^* \xleftarrow{} \text{argmin}(\alpha_{sim}L_{sim} + \alpha_{ent}L_{ent})
 \end{equation}

The training process (optimisation problem) results in a new $f_{emb,\theta_e^*}$ which embeds images so they can be classified easily, and a new $f_{cls,\theta_c^*}$ which can classify image embeddings into classes/clusters.

\begin{figure*}[h]
\centering

\begin{minipage}{0.32\linewidth}
    \centering
    \vspace*{-1.3cm}
    \includegraphics[width=\linewidth, height=5.2cm]{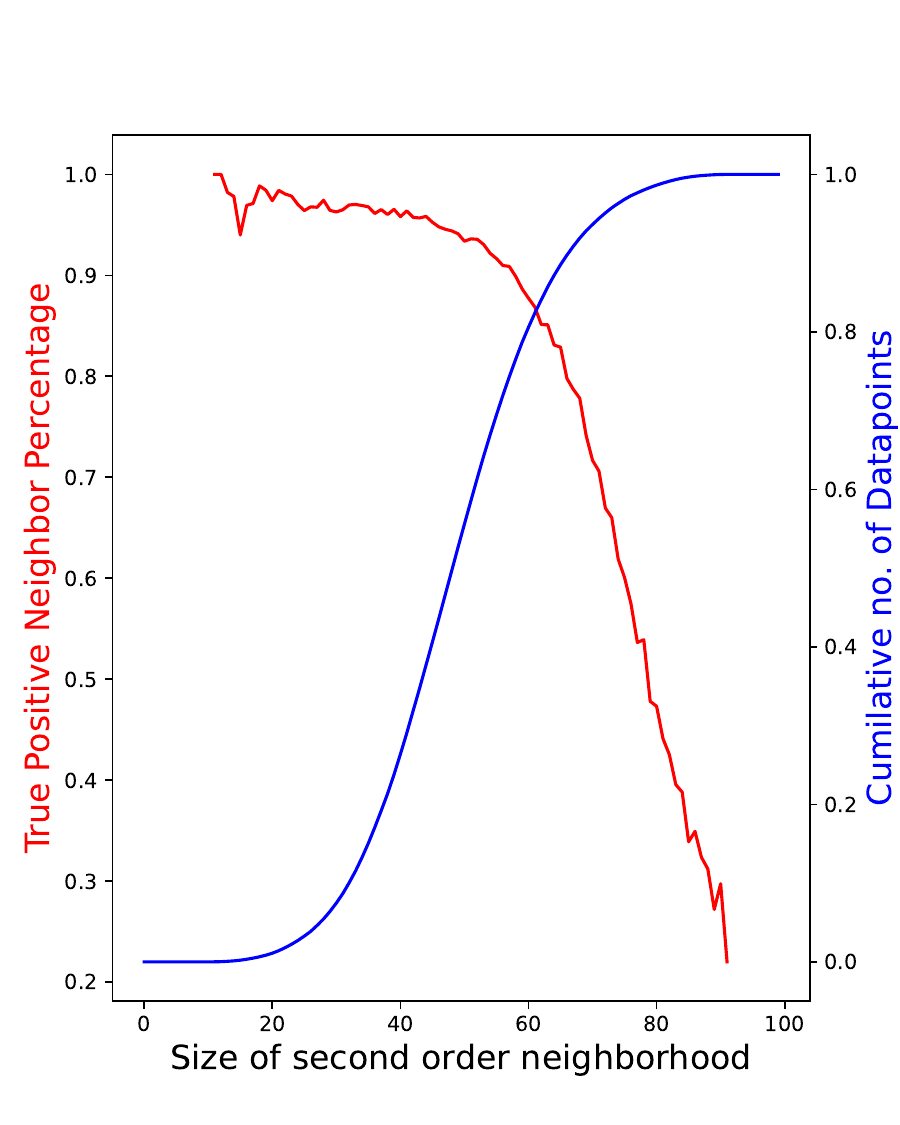}
    \caption{\textbf{True positives vs. second-order neighborhood size.} A threshold of 70 significantly reduces false positives (red) while conserving usable data points (blue).}
    \label{fig:sec_ord_neigh}
\end{minipage}
\hfill
\begin{minipage}{0.32\linewidth}
    \centering
    \vspace*{-\topskip}
    \includegraphics[width=\linewidth, height=5cm]{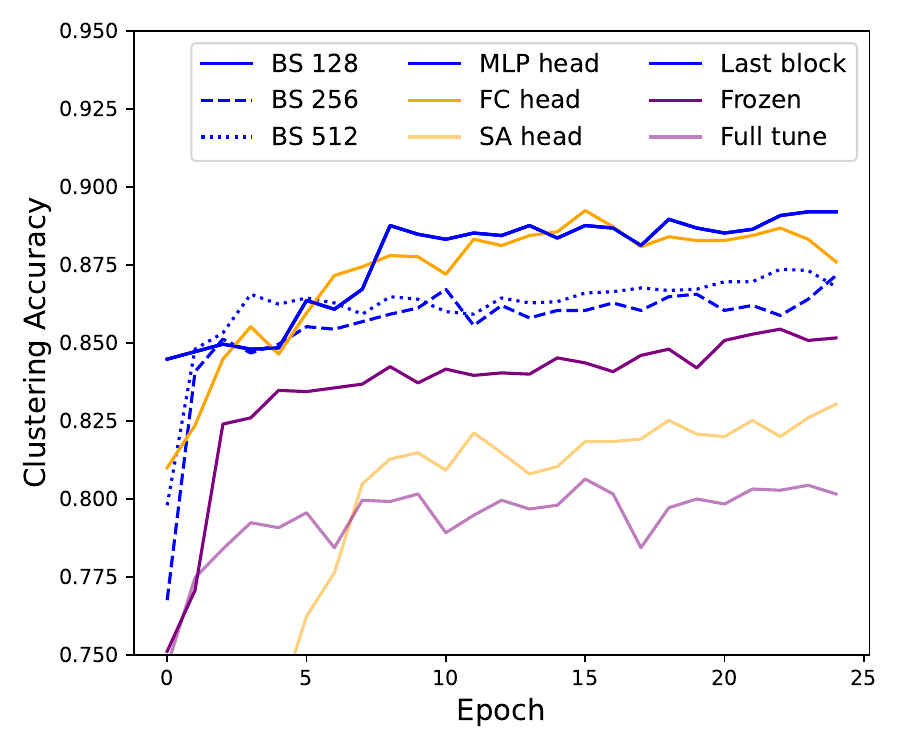}
    \caption{\textbf{Convergence behavior} for clustering on ImageNet-50. We compare batch sizes, finetuning levels (frozen, full-finetune, last block), and clustering heads (MLP, fully-connected, self-attention).}
    \label{fig:hyperparameter_convergence}
\end{minipage}
\hfill
\begin{minipage}{0.32\linewidth}
    \centering
    \vspace*{-\topskip}
    \includegraphics[width=\linewidth, height=5cm]{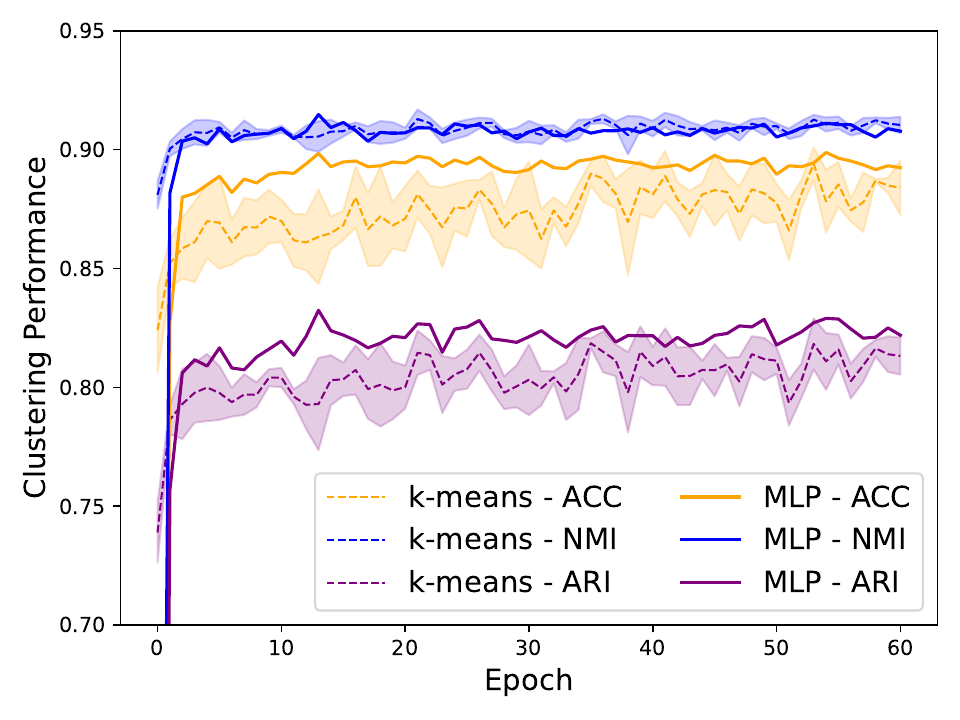}
    \caption{\textbf{k-Means accuracy over time.} k-Means results for ImageNet-50 using features extracted at different training epochs. The learned representations improve clustering as training progresses.}
    \label{fig:kmeans_over_time}
\end{minipage}

\end{figure*}

\begin{table}[t]
\centering

\resizebox{0.75\linewidth}{!}{
\begin{tabular}{@{}cccc c@{}}
    \toprule
    Positives & Negatives & Entropy & Contrastive & Accuracy \\
    \midrule
    \checkmark &  &  &  & 2.03 \\
    \checkmark & \checkmark &  &  & 17.64 \\
    \checkmark &  & \checkmark &  & {\ul 89.48} \\
    \checkmark & \checkmark & \checkmark &  & \textbf{90.12} \\
    \checkmark & \checkmark &  & \checkmark & 14.12 \\
    \checkmark & \checkmark & \checkmark & \checkmark & 88.24  \\
    \bottomrule
\end{tabular}
}
\caption{\textbf{Clustering ablations} for ImageNet-50 using negative neighbor mining, entropy loss, and contrastive loss with $\lambda_{POS}=1, \lambda_{NEG}=1, \lambda_{ENT}=3, \lambda_{CON}=5$ after 50 epochs.}
\label{tab:ablation_commponents}

\resizebox{0.75\linewidth}{!}{
\begin{tabular}{@{}cccc ccc@{}}
    \toprule
    \multicolumn{2}{c}{\textbf{Positive Neighbors}} & \multicolumn{2}{c}{\textbf{Negative Neighbors}} & \multicolumn{3}{c}{\textbf{Accuracy}} \\ 
    \cmidrule(l){1-2} \cmidrule(l){3-4} \cmidrule(l){5-7}
    \textbf{$D_L$} & \textbf{$D_U$} & \textbf{$D_L$} & \textbf{$D_U$} & \textbf{All} & \textbf{Old} & \textbf{New} \\ 
    \midrule
    Labeled & Mined & Random & Random & 82.22 & {\ul 92.36} & 77.13 \\
    Labeled & Cleaned & Mined & Mined & \textbf{83.22} & 91.97 & \textbf{78.82} \\
    Labeled & Cleaned & Labeled & Mined & {\ul 82.66} & \textbf{92.28} & {\ul 77.83} \\
    \bottomrule
\end{tabular}
}
\caption{\textbf{GCD ablations} (ImageNet-100) with different neighbor configurations after 50 epochs. Supplementing labeled positives and negatives with \textit{mined} negatives and \textit{cleaned} positives improves performance.}
\label{tab:ablation_gcd}

\end{table}  

\begin{figure}[h]
    \centering
    \includegraphics[width=0.75\linewidth]{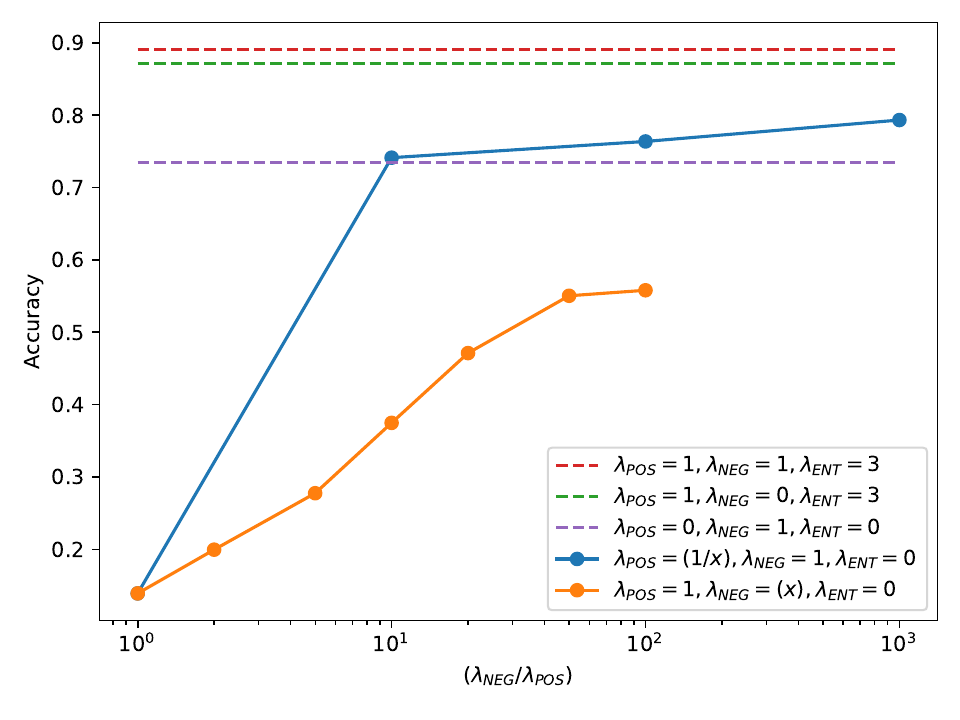}
    \caption{\textbf{Accuracy variation in ImageNet-50 clustering} with different $\lambda$ values.}
    \label{fig:lambda_trends}
\end{figure}

\subsection{Implementation details}

Vit B/16 backbone is used as $f_{emb}$. A two-layer MLP is used as $f_{cls}$. $\theta_e$ is initialized with DINO \citep{caron2021emerging} pretrained weights. It is further pretrained as per SimGCD \citep{Wen_2023_ICCV} for GCD experiments. $\theta_c$ is initialized at random. $x_i$ images undergo different transformations during training and testing times. They are always fed into the ViT as 3 colors $224\times224$ sized normalized tensors. Network parameters are optimized by Adam  with an initial learning rate of $10^{-4}$ which is cosine annealed for 100 epochs. Hyperparameters are given in the Appendix.

\section{Experiments}
\label{sec:results}

\subsection{Experiment Setup}

We evaluate our framework on datasets with 10, 50, 100, and 200 classes. We primarily stick with ImageNet-like datasets for their image quality and the coverage of a wide range of generic objects. It should be notes that we use a DINO ViT-B/16 backbone that is pretrained on ImageNet data without using any label information. This is crucial to have a fair comparison for unsupervised and semi-supervised tasks we perform.

We train and test UNIC on splits of ImageNet consisting of 50, 100 and 200 classes. 
In addition, we experiment on STL-10 dataset which consists of ImageNet-like images, but at lower resolution. 
For clustering, we do not utilize the labels of the training images. 
We follow the ImageNet-50, ImageNet-100 and ImageNet-200 splits from the recent literature to be consistent in comparison~\citep{TEMI_BMVC2023}. 
We use a different split for ImageNet-100 in GCD experiments to match the GCD literature~\citep{GCD_CVPR2022}. 
We utilize the labels of a fraction of images as per the GCD setting.
See Appendix A.2 for details of other datasets.

The proposed system is evaluated on the test splits of ImageNet and STL~\cite{pmlr-v15-coates11a} for clustering using the target labels. 
The evaluation for GCD is done with the target labels of the unlabelled portion of the training set. 
While the most early work in GCD has utilized the labelled test split of the dataset as an early stopping strategy~\citep{Wen_2023_ICCV}, we exclude this to maintain a realistic open world setting.
We conduct all experiments on a Nvidia RTX4000 GPU with 16GB of vRAM.

\begin{figure*}[h]
    \centering

    \begin{subfigure}[b]{0.32\textwidth}
    \centering
    \includegraphics[width=\textwidth]{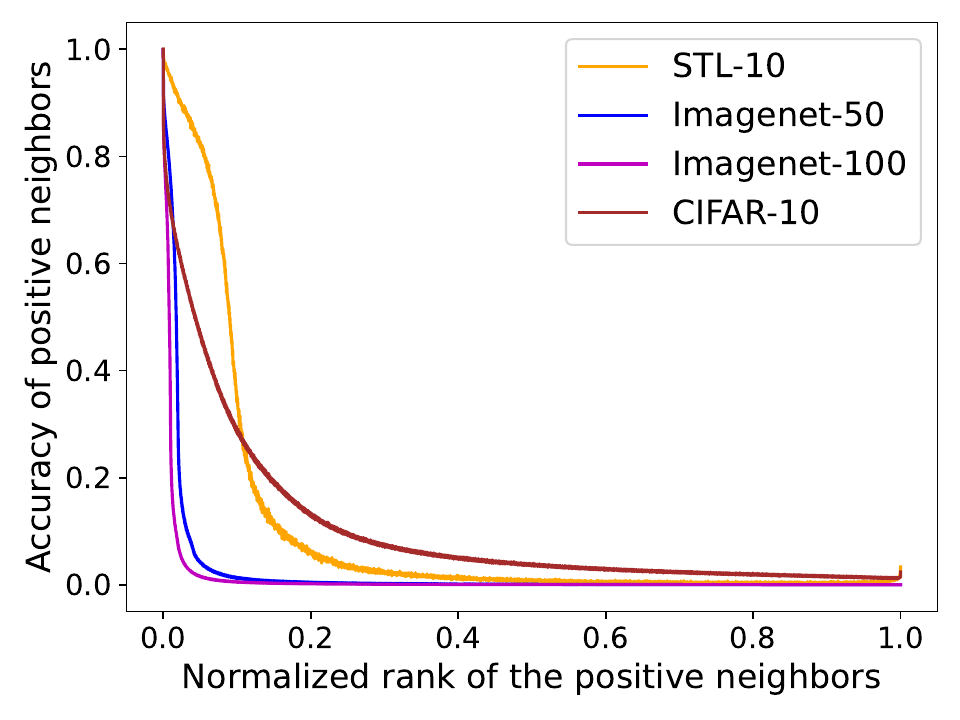}
    \end{subfigure}
    \hfill
    \begin{subfigure}[b]{0.32\textwidth}
    \centering
    \includegraphics[width=\textwidth]{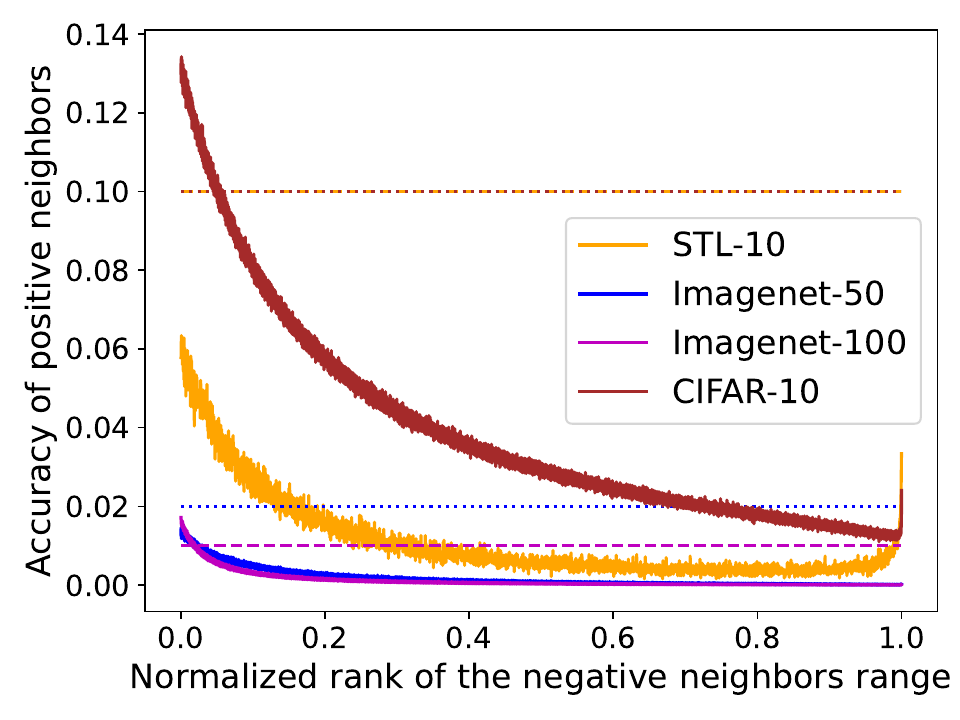}
    \end{subfigure}
    \hfill
    \begin{subfigure}[b]{0.32\textwidth}
    \centering
    \includegraphics[width=\textwidth]{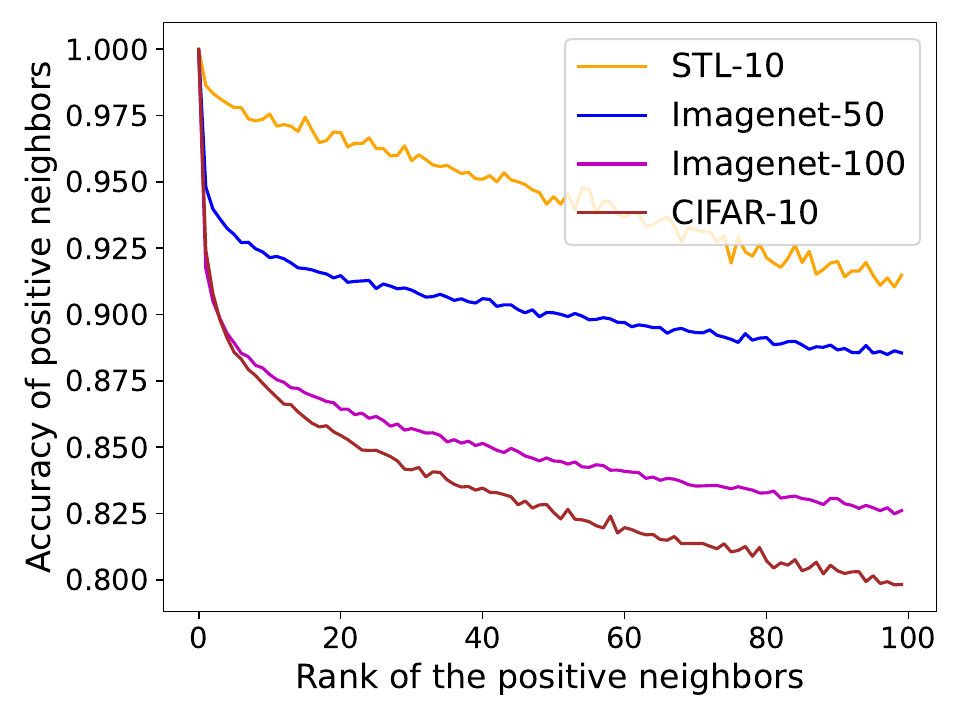}
    \end{subfigure}
    
    \caption{\textbf{Neighbor Analysis.} We show the amount of times a potential neighbor belongs to the same class as an anchor image (``Neighbor Accuracy''), depending on their relative distance, with the normalized index of the potential matching image in a sorted list of all images plotted on the x-axis (left). Notice there is a certain point where the amount of matches is comfortably low -- we mine negative neighbors from this range. We zoom in on these negative neighbors, and compare the accuracy (which we want to be low) to random sampling, represented by the horizontal dashed lines (middle). We also zoom in on the positive neighbor range (right).\label{fig:negative_neighbor_threshold}}
    
\end{figure*}

\begin{figure*}[h]
\centering

    \begin{minipage}{0.32\linewidth}

    \centering
    \includegraphics[width=0.9\textwidth]{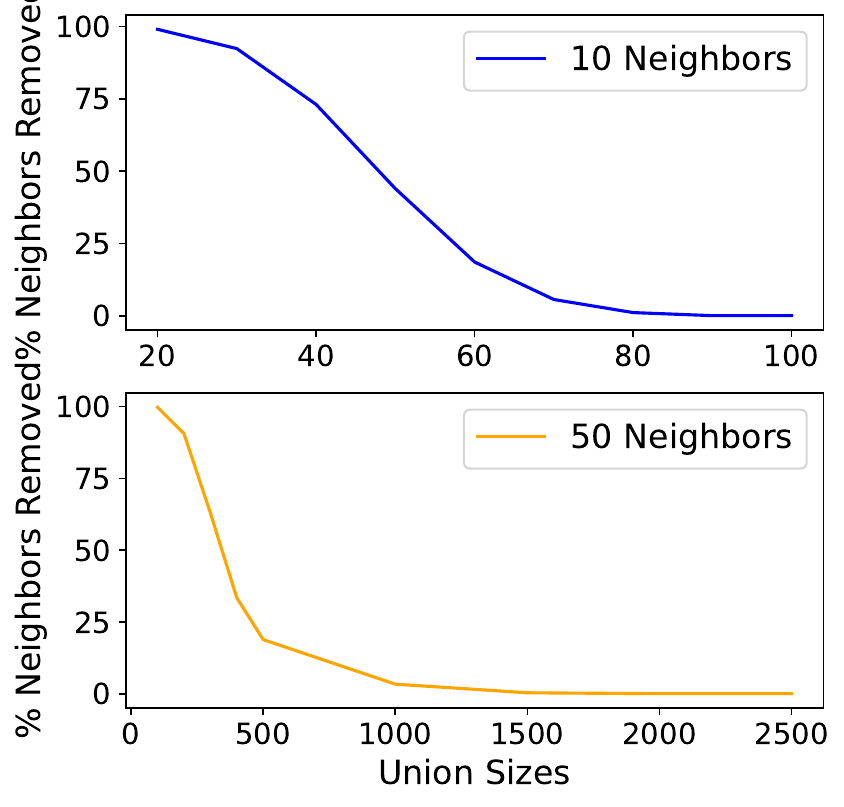}
    \caption{\textbf{Neighbors Removed by Union Size} for $\tau_1 = 10, 50$ on ImageNet-50.\label{fig:union_removal}}

\end{minipage}
\hfill
\begin{minipage}{0.32\linewidth}

    \centering
    \includegraphics[width=0.9\textwidth]{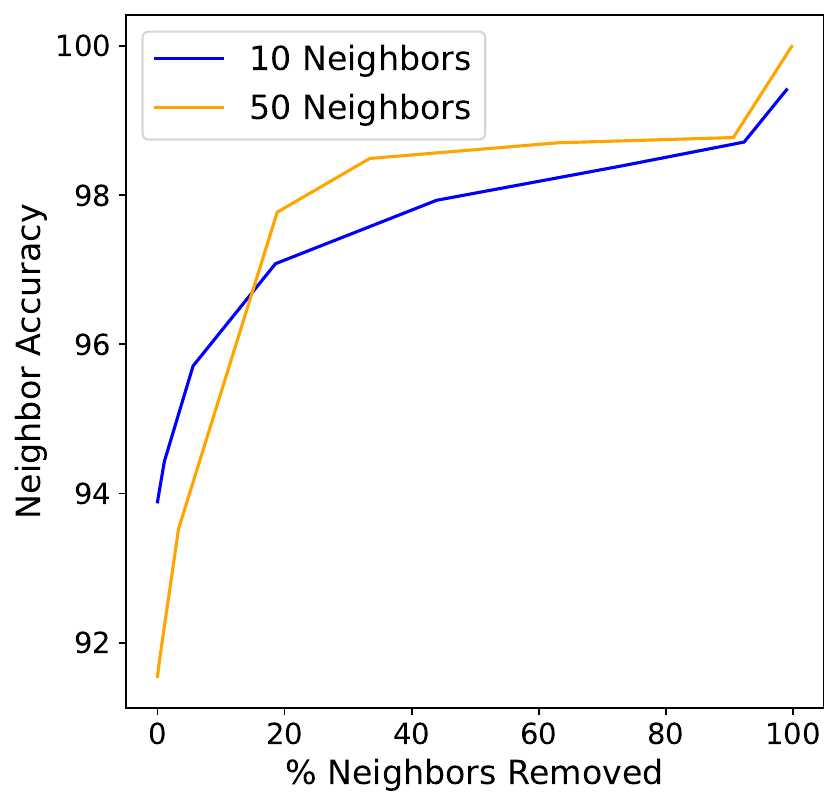}
    \caption{\textbf{Neighbor Purity by Samples Removed} for $\tau_1 = 10, 50$ on ImageNet-50.\label{fig:union_neighbor_purity}}

\end{minipage}
\hfill
\begin{minipage}{0.32\linewidth}

    \centering
    \includegraphics[width=0.9\textwidth]{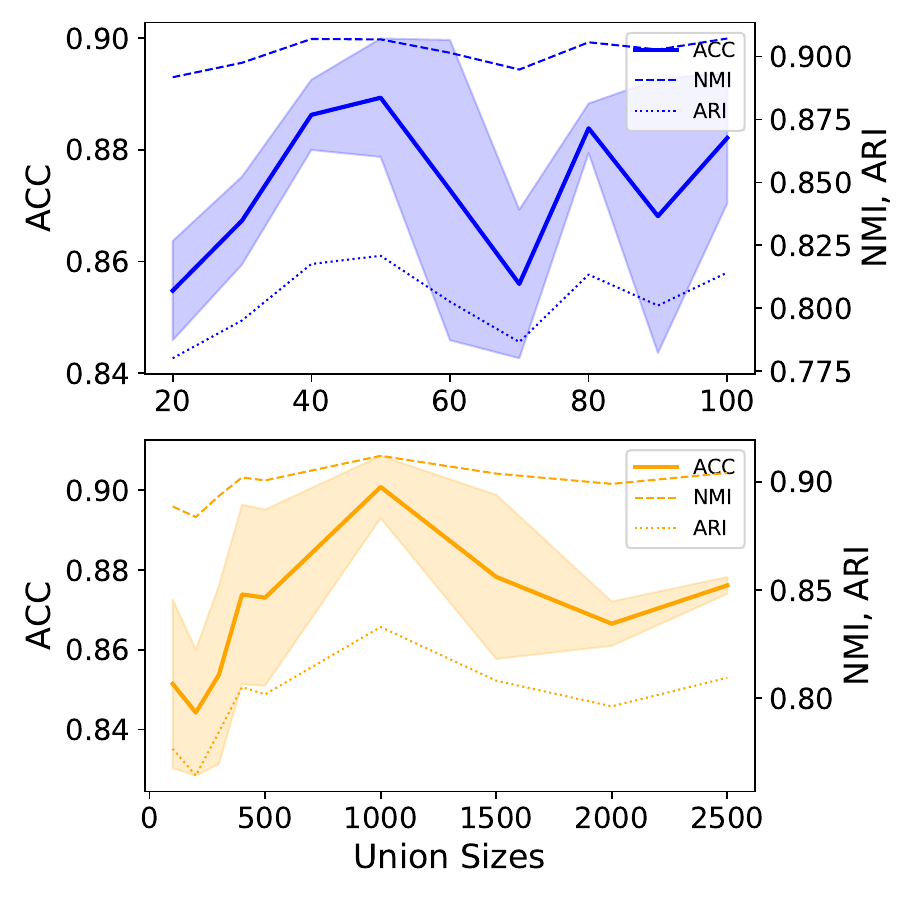}
    \caption{\textbf{Results by Union Size} for $\tau_1 = 10, 50$ (top,bottom) on ImageNet-50.\label{fig:union_clustering_performance}}

\end{minipage}
\end{figure*}

\subsection{Results}

First, we provide results for clustering where we largely outperforms the prior work for all metrics on the ImageNet splits while being competitive on CIFAR~\cite{krizhevsky2009learning}.
Notably, while the prior work does not outperform properly-tuned k-means consistently (see TEMI on ImageNet-50, for example), ours does in all cases. 
Also, since we finetune the last block of the backbone, we can even improve the k-means results (see Figure~\ref{fig:kmeans_over_time}).
We investigate this phenomenon further in Section~\ref{subsec:analysis}.

Next, UNIC achieves SOTA on generic and fine grained GCD datasets as per Table~\ref{tab:gcd_results}.
Furthermore, for the more realistic (higher resolution images, more classes) dataset, we achieve SOTA performance on the unlabelled data (``All Classes''), primarily because of our methods strong performance on ``Old Classes'' (unlabelled images from seen classes).
Our model's reliance on neighbor quality can sometimes hurt performance.
However, in the case of GCD it is clearly a major benefit, since we have perfect neighbors for half of the images for ``Old Classes.''
We examine our reliance on neighbor quality further in Section~\ref{subsec:analysis}.

\subsection{Ablations}

We conduct ablations to understand three items. Firstly, we validate our hypothesis about second-order neighborhood sizes being a proxy for the percentage of true positives in the first-order neighborhood. This analysis is present in Figure~\ref{fig:sec_ord_neigh}. It shows that we are able to pick a threshold for second-order neighborhood size (a point in x axis) such that we mostly get true positive neighbors (y value of the red line) for a large enough fraction of the dataset (y value of the blue line). More information and visualization for the phenomena is given in Appendix A.3.

Secondly, we examine the utility of the multiple loss terms proposed in our pipeline ($L_{pos},L_{neg},L_{ent}$). The results are given in Figure~\ref{fig:lambda_trends}. The dotted red line above the dotted green line shows the gain from the proposed $L_{neg}$ term. More importantly, the other three lines show how $L_{neg}$ by itself is enough to converge to a stable solution. This is the first success in clustering literature to get rid of the $L_{ent}$ term. In addition, the trends show that we can further improve the performance by the choice of $L_{neg}/L_{pos}$. This is further demonstrated with the results in Table~\ref{tab:ablation_commponents}. In addition, we also find that using a contrastive loss, as in some prior works, actually makes our clustering slightly worse. This can be attributed to the better supervision signal coming out of the proposed clustering head compared to the signal coming from the DINO contrastive head.

\begin{figure*}[t!]
    \centering
    \begin{tabular}{ccc}
      \includegraphics[width=0.49\linewidth]{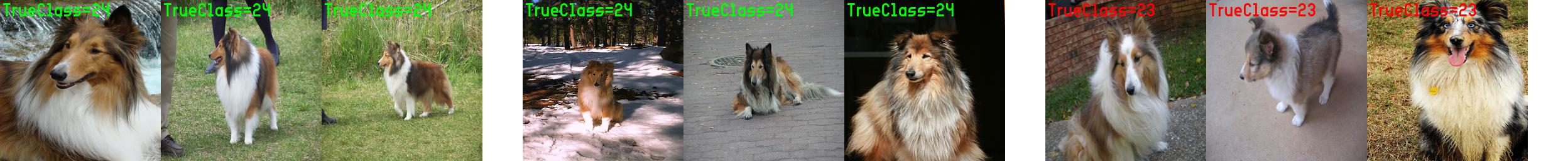}  & \includegraphics[width=0.49\linewidth]{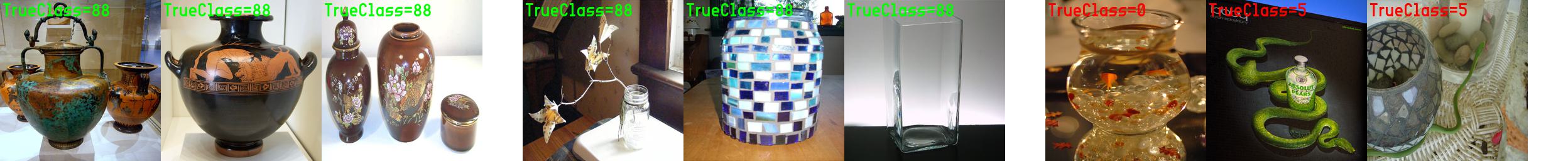} \\
      \includegraphics[width=0.49\linewidth]{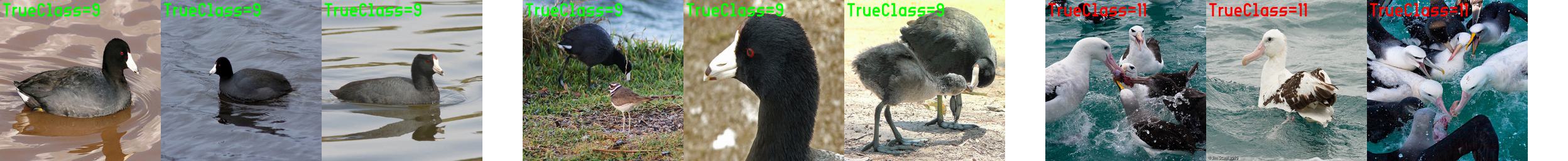}  & \includegraphics[width=0.49\linewidth]{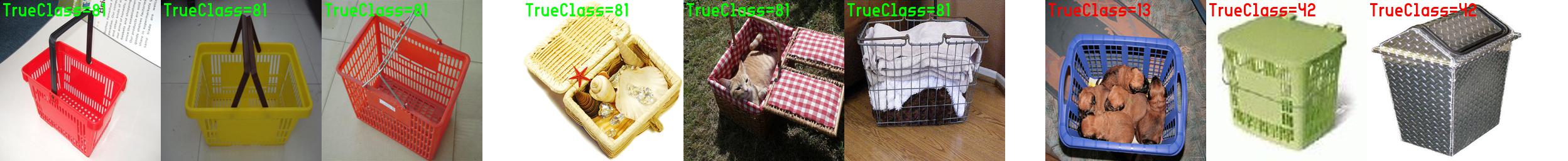} \\
      \includegraphics[width=0.49\linewidth]{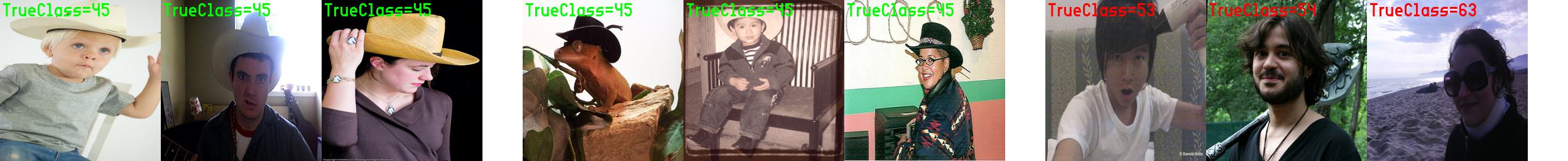}  & \includegraphics[width=0.49\linewidth]{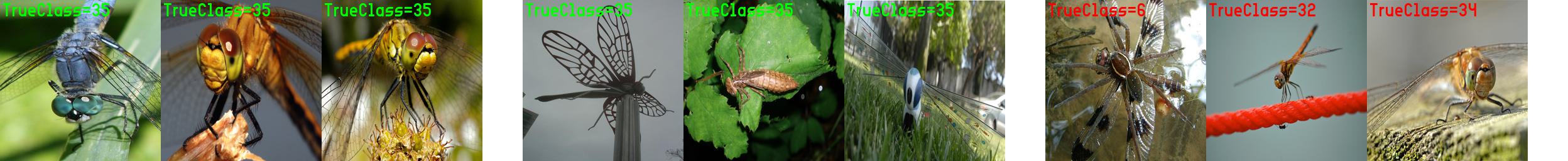} \\
      \includegraphics[width=0.49\linewidth]{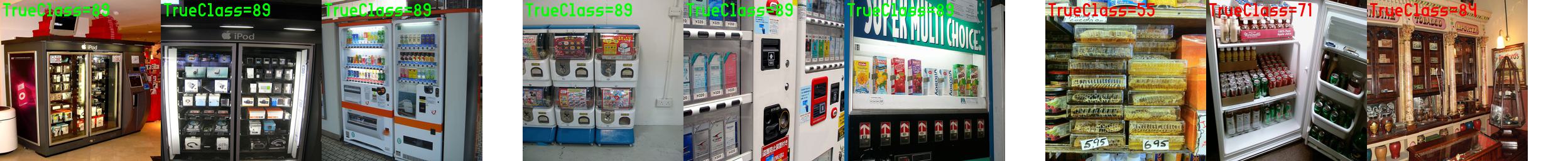}  & \includegraphics[width=0.49\linewidth]{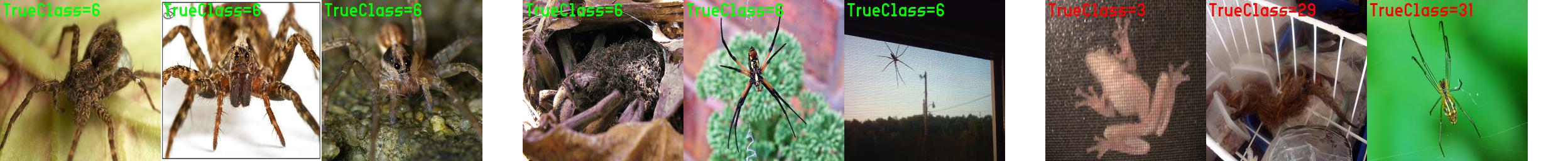} \\
      (a) Clustering Examples for ImageNet-100 & (b) GCD Examples for ImageNet-100 \\
\end{tabular}

    \vspace{-0.5cm}
    \caption{\textbf{Cluster examples.} Each row corresponds to a cluster, for which we show triplets: (left) the highest confidence true positives, (middle) the lowest confidence true positives, and (right) randomly-sampled false positives. We find that in both cases UNIC is able to identify salient visual similarities, even when these are challenging, such as the lizard in the cowboy hat on the $\text{3}^\text{rd}$ row (a).\label{fig:sample-results}}
    
\end{figure*}

Thirdly, we conduct ablation studies to  further analyze the effect of neighbors and their labelling on performance, we compare the impacts of random, mined, cleaned, and ground truth (labeled) neighbors for the two sets of images in GCD (labelled and unlabelled).
In Table~\ref{tab:ablation_gcd}, we find that mining and cleaning positives for new classes improves upon the result we obtain by using the ground truth positives for old classes. 
In addition, we see improvement by mining negatives as per our algorithm, as opposed to using only ground truth neighbors or random neighbors. The random negative neighbors perform worse due to the false negatives in the picked sets as per Figure~\ref{fig:negative_neighbor_threshold}. Picking labeled negatives for $D_L$ causes an over-representation of negatives from old classes for these anchor images. This is further explained in Appendix A.11.%

\subsection{Analysis}
\label{subsec:analysis}

We first show how our method converges, depending on the selection of certain key hyperparameters, in Figure~\ref{fig:hyperparameter_convergence}.
We show that we carefully choose our batch size, clustering head design, and level of finetuning to optimize performance.
As mentioned previously, since we partially finetune the ViT backbone, its learned representation quality improves. 
We measure this improvement by tracking the k-means performance across training time in Figure~\ref{fig:kmeans_over_time}.
So, we conclude that while we target these classification tasks (GCD and clustering) directly, our method is indeed in some forms a useful strategy for unsupervised representation learning, at least as a finetuning method.

To select our threshold for negative neighbors, we consider the set of images.
For each image, consider its distance from all other images, and sort them accordingly.
For every image in the dataset, we thus have a sorted list of neighbor images, and each list is of the same length (containing all images in the dataset except the anchor).
Now for each index we count how often the image belongs to the same class as the anchor, and plot this in Figure~\ref{fig:negative_neighbor_threshold}.
This reveals how we mine our negative neighbors.
Contrary to positive mining, we are not limited to some $k$ neighbors -- instead, we can mine more than half of all images as negatives, for any given anchor. Figure ~\ref{fig:negative_neighbor_threshold} (middle) validates our heurestical choices for $\tau_2$ because the false negatives are much lower than the dotted lines for any sufficient enough $\tau_2$. It should be noted that we do not tune $\tau_2$ extensively inorder to stay true for the problem settings.
We show the reliability of the positive neighbors in the Figure~\ref{fig:negative_neighbor_threshold} (right), giving some intuition as why we tend to set $10 \leq \tau_1 \leq 50$.

For cleaning, we show how we remove more neighbors as we increase the union sizes in Figure~\ref{fig:union_removal}.
Then, we notice that for smaller unions, our positive neighbors tend to be more ``pure'' in Figure~\ref{fig:union_neighbor_purity}.
That is, they frequently belong to the same class, and it is more rare that positive neighbors belong to different classes.
As expected, when we use these more pure positive neighbors, we get better clustering performance, which we show in Figure~\ref{fig:union_clustering_performance}.
This holds across different numbers of neighbors.

We find that UNIC learns sensible clusters both for clustering and GCD in Figure~\ref{fig:sample-results}. 
It tends to group things along the intended class boundaries, even when samples are strange, such as the instance with the lizard wearing the cowboy hat.
Most false positives are visually similar to the true positives.

\section{Conclusion}
\label{sec:conclusion}

In this paper, we present an approach that improves on state-of-the-art for image clustering and GCD.
Due to its flexibility and general formulation, our approach also conceptually unifies these two related vision problems.
We also provide robust analysis of our major contributions, namely neighbor cleaning, negative neighbor mining, and other relevant hyperparameters.
We hope our work here serves to break down the barriers between different levels of supervision, as we work towards real-world solutions that use labels in efficient, intuitive ways while still working well when labels are totally unavailable.

\textbf{Limitations:} 
Our proposed UNIC leverages neighbor information very well.
However, it is sensitive to very high levels of noise in neighbor mining.

\textbf{Future Work:}
Our work has shown that deep clustering does not require an entropy maximization objective. Therefore, future work can relax the assumption on equally distributed classes and attempt to solve unbalanced clustering problems with UNIC (at $\lambda_{ENT}=0$). A natural extension of this work is to apply UNIC's losses can be applied to open world object detection.

\textbf{Broader Impacts:} 
Image recognition can have negative impacts with uses in surveillance and autonomous weapons.
Alternatively, good image clustering can help with medical imaging, where labels are scarce, for tracking efforts with endangered species, and other such applications.

\FloatBarrier
\clearpage

{
    \small
    \bibliographystyle{unsrtnat}
    \bibliography{ref}

\begin{thebibliography}{70}
\providecommand{\natexlab}[1]{#1}
\providecommand{\url}[1]{\texttt{#1}}
\expandafter\ifx\csname urlstyle\endcsname\relax
  \providecommand{\doi}[1]{doi: #1}\else
  \providecommand{\doi}{doi: \begingroup \urlstyle{rm}\Url}\fi

\bibitem[Deng et~al.(2009)Deng, Dong, Socher, Li, Li, and Fei-Fei]{5206848}
Jia Deng, Wei Dong, Richard Socher, Li-Jia Li, Kai Li, and Li~Fei-Fei.
\newblock Imagenet: A large-scale hierarchical image database.
\newblock In \emph{2009 IEEE Conference on Computer Vision and Pattern Recognition}, pages 248--255, 2009.
\newblock \doi{10.1109/CVPR.2009.5206848}.

\bibitem[Koch et~al.(2015)Koch, Zemel, Salakhutdinov, et~al.]{koch2015siamese}
Gregory Koch, Richard Zemel, Ruslan Salakhutdinov, et~al.
\newblock Siamese neural networks for one-shot image recognition.
\newblock In \emph{ICML deep learning workshop}, volume~2. Lille, 2015.

\bibitem[Bendale and Boult(2015)]{Bendale_2015}
Abhijit Bendale and Terrance Boult.
\newblock Towards open world recognition.
\newblock In \emph{2015 IEEE Conference on Computer Vision and Pattern Recognition (CVPR)}. IEEE, June 2015.
\newblock \doi{10.1109/cvpr.2015.7298799}.
\newblock URL \url{http://dx.doi.org/10.1109/CVPR.2015.7298799}.

\bibitem[Snell et~al.(2017)Snell, Swersky, and Zemel]{snell2017prototypical}
Jake Snell, Kevin Swersky, and Richard~S. Zemel.
\newblock Prototypical networks for few-shot learning, 2017.

\bibitem[Vinyals et~al.(2017)Vinyals, Blundell, Lillicrap, Kavukcuoglu, and Wierstra]{vinyals2017matching}
Oriol Vinyals, Charles Blundell, Timothy Lillicrap, Koray Kavukcuoglu, and Daan Wierstra.
\newblock Matching networks for one shot learning, 2017.

\bibitem[Suri et~al.(2023)Suri, Rambhatla, Chellappa, and Shrivastava]{suri2023sparsedet}
Saksham Suri, Sai~Saketh Rambhatla, Rama Chellappa, and Abhinav Shrivastava.
\newblock Sparsedet: Improving sparsely annotated object detection with pseudo-positive mining, 2023.

\bibitem[Gansbeke et~al.(2020)Gansbeke, Vandenhende, Georgoulis, Proesmans, and Van~Gool]{SCAN_ECCV2020}
Wouter~Van Gansbeke, Simon Vandenhende, Stamatios Georgoulis, Marc Proesmans, and Luc Van~Gool.
\newblock Scan: Learning to classify images without labels.
\newblock In \emph{Proceedings of the European Conference on Computer Vision}, 2020.

\bibitem[Vaze et~al.(2022)Vaze, Han, Vedaldi, and Zisserman]{GCD_CVPR2022}
Sagar Vaze, Kai Han, Andrea Vedaldi, and Andrew Zisserman.
\newblock Generalized category discovery.
\newblock In \emph{{IEEE/CVF} Conference on Computer Vision and Pattern Recognition, {CVPR} 2022, New Orleans, LA, USA, June 18-24, 2022}, pages 7482--7491. {IEEE}, 2022.
\newblock \doi{10.1109/CVPR52688.2022.00734}.
\newblock URL \url{https://doi.org/10.1109/CVPR52688.2022.00734}.

\bibitem[Caron et~al.(2021)Caron, Touvron, Misra, J{\'e}gou, Mairal, Bojanowski, and Joulin]{caron2021emerging}
Mathilde Caron, Hugo Touvron, Ishan Misra, Herv{\'e} J{\'e}gou, Julien Mairal, Piotr Bojanowski, and Armand Joulin.
\newblock Emerging properties in self-supervised vision transformers.
\newblock \emph{arXiv preprint arXiv:2104.14294}, 2021.

\bibitem[Dosovitskiy et~al.(2021)Dosovitskiy, Beyer, Kolesnikov, Weissenborn, Zhai, Unterthiner, Dehghani, Minderer, Heigold, Gelly, Uszkoreit, and Houlsby]{dosovitskiy2021image}
Alexey Dosovitskiy, Lucas Beyer, Alexander Kolesnikov, Dirk Weissenborn, Xiaohua Zhai, Thomas Unterthiner, Mostafa Dehghani, Matthias Minderer, Georg Heigold, Sylvain Gelly, Jakob Uszkoreit, and Neil Houlsby.
\newblock An image is worth 16x16 words: Transformers for image recognition at scale, 2021.

\bibitem[Pu et~al.(2023)Pu, Zhong, and Sebe]{DCCL_CVPR2023}
Nan Pu, Zhun Zhong, and Nicu Sebe.
\newblock Dynamic conceptional contrastive learning for generalized category discovery.
\newblock In \emph{{IEEE/CVF} Conference on Computer Vision and Pattern Recognition, {CVPR} 2023, Vancouver, BC, Canada, June 17-24, 2023}, pages 7579--7588. {IEEE}, 2023.
\newblock \doi{10.1109/CVPR52729.2023.00732}.
\newblock URL \url{https://doi.org/10.1109/CVPR52729.2023.00732}.

\bibitem[Doersch et~al.(2015)Doersch, Gupta, and Efros]{doersch2015unsupervised}
Carl Doersch, Abhinav Gupta, and Alexei~A Efros.
\newblock Unsupervised visual representation learning by context prediction.
\newblock In \emph{Proceedings of the IEEE international conference on computer vision}, pages 1422--1430, 2015.

\bibitem[Gupta et~al.(2020)Gupta, Singh, and Shrivastava]{gupta2020patchvae}
Kamal Gupta, Saurabh Singh, and Abhinav Shrivastava.
\newblock Patchvae: Learning local latent codes for recognition, 2020.

\bibitem[Noroozi and Favaro(2016)]{noroozi2016unsupervised}
Mehdi Noroozi and Paolo Favaro.
\newblock Unsupervised learning of visual representations by solving jigsaw puzzles.
\newblock In \emph{European conference on computer vision}, pages 69--84. Springer, 2016.

\bibitem[Larsson et~al.(2016)Larsson, Maire, and Shakhnarovich]{larsson2016learning}
Gustav Larsson, Michael Maire, and Gregory Shakhnarovich.
\newblock Learning representations for automatic colorization.
\newblock In \emph{European conference on computer vision}, pages 577--593. Springer, 2016.

\bibitem[Zhang et~al.(2016)Zhang, Isola, and Efros]{zhang2016colorful}
Richard Zhang, Phillip Isola, and Alexei~A Efros.
\newblock Colorful image colorization.
\newblock In \emph{European conference on computer vision}, pages 649--666. Springer, 2016.

\bibitem[Pathak et~al.(2016)Pathak, Krahenbuhl, Donahue, Darrell, and Efros]{pathak2016context}
Deepak Pathak, Philipp Krahenbuhl, Jeff Donahue, Trevor Darrell, and Alexei~A Efros.
\newblock Context encoders: Feature learning by inpainting.
\newblock In \emph{Proceedings of the IEEE conference on computer vision and pattern recognition}, pages 2536--2544, 2016.

\bibitem[Gidaris et~al.(2018)Gidaris, Singh, and Komodakis]{gidaris2018unsupervised}
Spyros Gidaris, Praveer Singh, and Nikos Komodakis.
\newblock Unsupervised representation learning by predicting image rotations.
\newblock \emph{arXiv preprint arXiv:1803.07728}, 2018.

\bibitem[Donahue and Simonyan(2019)]{donahue2019large}
Jeff Donahue and Karen Simonyan.
\newblock Large scale adversarial representation learning, 2019.

\bibitem[Doersch and Zisserman(2017)]{doersch2017multi}
Carl Doersch and Andrew Zisserman.
\newblock Multi-task self-supervised visual learning.
\newblock In \emph{Proceedings of the IEEE International Conference on Computer Vision}, pages 2051--2060, 2017.

\bibitem[Wu et~al.(2018)Wu, Xiong, Yu, and Lin]{Wu_2018_CVPR}
Zhirong Wu, Yuanjun Xiong, Stella~X. Yu, and Dahua Lin.
\newblock Unsupervised feature learning via non-parametric instance discrimination.
\newblock In \emph{Proceedings of the IEEE Conference on Computer Vision and Pattern Recognition (CVPR)}, June 2018.

\bibitem[Chen et~al.(2020{\natexlab{a}})Chen, Kornblith, Norouzi, and Hinton]{pmlr-v119-chen20j}
Ting Chen, Simon Kornblith, Mohammad Norouzi, and Geoffrey Hinton.
\newblock A simple framework for contrastive learning of visual representations.
\newblock In Hal~Daumé III and Aarti Singh, editors, \emph{Proceedings of the 37th International Conference on Machine Learning}, volume 119 of \emph{Proceedings of Machine Learning Research}, pages 1597--1607. PMLR, 13--18 Jul 2020{\natexlab{a}}.
\newblock URL \url{https://proceedings.mlr.press/v119/chen20j.html}.

\bibitem[Chen et~al.(2020{\natexlab{b}})Chen, Fan, Girshick, and He]{chen2020improved}
Xinlei Chen, Haoqi Fan, Ross Girshick, and Kaiming He.
\newblock Improved baselines with momentum contrastive learning.
\newblock \emph{arXiv preprint arXiv:2003.04297}, 2020{\natexlab{b}}.

\bibitem[Misra and Maaten(2020)]{misra2020self}
Ishan Misra and Laurens van~der Maaten.
\newblock Self-supervised learning of pretext-invariant representations.
\newblock In \emph{Proceedings of the IEEE/CVF Conference on Computer Vision and Pattern Recognition}, pages 6707--6717, 2020.

\bibitem[Bachman et~al.(2019)Bachman, Hjelm, and Buchwalter]{bachman2019learning}
Philip Bachman, R~Devon Hjelm, and William Buchwalter.
\newblock Learning representations by maximizing mutual information across views.
\newblock \emph{arXiv preprint arXiv:1906.00910}, 2019.

\bibitem[Chen et~al.(2020{\natexlab{c}})Chen, Kornblith, Norouzi, and Hinton]{chen2020simple}
Ting Chen, Simon Kornblith, Mohammad Norouzi, and Geoffrey Hinton.
\newblock A simple framework for contrastive learning of visual representations.
\newblock In \emph{International conference on machine learning}, pages 1597--1607. PMLR, 2020{\natexlab{c}}.

\bibitem[He et~al.(2019)He, Fan, Wu, Xie, and Girshick]{DBLP:journals/corr/abs-1911-05722}
Kaiming He, Haoqi Fan, Yuxin Wu, Saining Xie, and Ross~B. Girshick.
\newblock Momentum contrast for unsupervised visual representation learning.
\newblock \emph{CoRR}, abs/1911.05722, 2019.
\newblock URL \url{http://arxiv.org/abs/1911.05722}.

\bibitem[Hjelm et~al.(2018)Hjelm, Fedorov, Lavoie-Marchildon, Grewal, Bachman, Trischler, and Bengio]{hjelm2018learning}
R~Devon Hjelm, Alex Fedorov, Samuel Lavoie-Marchildon, Karan Grewal, Phil Bachman, Adam Trischler, and Yoshua Bengio.
\newblock Learning deep representations by mutual information estimation and maximization.
\newblock \emph{arXiv preprint arXiv:1808.06670}, 2018.

\bibitem[Henaff(2020)]{henaff2020data}
Olivier Henaff.
\newblock Data-efficient image recognition with contrastive predictive coding.
\newblock In \emph{International Conference on Machine Learning}, pages 4182--4192. PMLR, 2020.

\bibitem[Oord et~al.(2018)Oord, Li, and Vinyals]{oord2018representation}
Aaron van~den Oord, Yazhe Li, and Oriol Vinyals.
\newblock Representation learning with contrastive predictive coding.
\newblock \emph{arXiv preprint arXiv:1807.03748}, 2018.

\bibitem[Tian et~al.(2020)Tian, Krishnan, and Isola]{tian2020contrastive}
Yonglong Tian, Dilip Krishnan, and Phillip Isola.
\newblock Contrastive multiview coding.
\newblock In \emph{Computer Vision--ECCV 2020: 16th European Conference, Glasgow, UK, August 23--28, 2020, Proceedings, Part XI 16}, pages 776--794. Springer, 2020.

\bibitem[Ye et~al.(2019)Ye, Zhang, Yuen, and Chang]{Ye_2019_CVPR}
Mang Ye, Xu~Zhang, Pong~C. Yuen, and Shih-Fu Chang.
\newblock Unsupervised embedding learning via invariant and spreading instance feature.
\newblock In \emph{Proceedings of the IEEE/CVF Conference on Computer Vision and Pattern Recognition (CVPR)}, June 2019.

\bibitem[Zbontar et~al.(2021)Zbontar, Jing, Misra, LeCun, and Deny]{DBLP:journals/corr/abs-2103-03230}
Jure Zbontar, Li~Jing, Ishan Misra, Yann LeCun, and St{\'{e}}phane Deny.
\newblock Barlow twins: Self-supervised learning via redundancy reduction.
\newblock \emph{CoRR}, abs/2103.03230, 2021.
\newblock URL \url{https://arxiv.org/abs/2103.03230}.

\bibitem[Chen and He(2021)]{chen2021exploring}
Xinlei Chen and Kaiming He.
\newblock Exploring simple siamese representation learning.
\newblock In \emph{Proceedings of the IEEE/CVF Conference on Computer Vision and Pattern Recognition}, pages 15750--15758, 2021.

\bibitem[Caron et~al.(2018)Caron, Bojanowski, Joulin, and Douze]{caron2018deep}
Mathilde Caron, Piotr Bojanowski, Armand Joulin, and Matthijs Douze.
\newblock Deep clustering for unsupervised learning of visual features.
\newblock In \emph{Proceedings of the European Conference on Computer Vision (ECCV)}, pages 132--149, 2018.

\bibitem[Asano et~al.(2019)Asano, Rupprecht, and Vedaldi]{asano2019self}
Yuki~Markus Asano, Christian Rupprecht, and Andrea Vedaldi.
\newblock Self-labelling via simultaneous clustering and representation learning.
\newblock \emph{arXiv preprint arXiv:1911.05371}, 2019.

\bibitem[Caron et~al.(2019)Caron, Bojanowski, Mairal, and Joulin]{caron2019unsupervised}
Mathilde Caron, Piotr Bojanowski, Julien Mairal, and Armand Joulin.
\newblock Unsupervised pre-training of image features on non-curated data.
\newblock In \emph{Proceedings of the IEEE/CVF International Conference on Computer Vision}, pages 2959--2968, 2019.

\bibitem[Caron et~al.(2020)Caron, Misra, Mairal, Goyal, Bojanowski, and Joulin]{caron2020unsupervised}
Mathilde Caron, Ishan Misra, Julien Mairal, Priya Goyal, Piotr Bojanowski, and Armand Joulin.
\newblock Unsupervised learning of visual features by contrasting cluster assignments.
\newblock \emph{arXiv preprint arXiv:2006.09882}, 2020.

\bibitem[Li et~al.(2021)Li, Zhou, Xiong, and Hoi]{li2021prototypical}
Junnan Li, Pan Zhou, Caiming Xiong, and Steven C.~H. Hoi.
\newblock Prototypical contrastive learning of unsupervised representations, 2021.

\bibitem[He et~al.(2021)He, Chen, Xie, Li, Doll{\'{a}}r, and Girshick]{DBLP:journals/corr/abs-2111-06377}
Kaiming He, Xinlei Chen, Saining Xie, Yanghao Li, Piotr Doll{\'{a}}r, and Ross~B. Girshick.
\newblock Masked autoencoders are scalable vision learners.
\newblock \emph{CoRR}, abs/2111.06377, 2021.
\newblock URL \url{https://arxiv.org/abs/2111.06377}.

\bibitem[Assran et~al.(2022)Assran, Caron, Misra, Bojanowski, Bordes, Vincent, Joulin, Rabbat, and Ballas]{assran2022masked}
Mahmoud Assran, Mathilde Caron, Ishan Misra, Piotr Bojanowski, Florian Bordes, Pascal Vincent, Armand Joulin, Michael Rabbat, and Nicolas Ballas.
\newblock Masked siamese networks for label-efficient learning, 2022.

\bibitem[Zhou et~al.(2022)Zhou, Wei, Wang, Shen, Xie, Yuille, and Kong]{zhou2022ibot}
Jinghao Zhou, Chen Wei, Huiyu Wang, Wei Shen, Cihang Xie, Alan Yuille, and Tao Kong.
\newblock ibot: Image bert pre-training with online tokenizer, 2022.

\bibitem[Huang et~al.(2022)Huang, Jin, Lu, Hou, Cheng, Fu, Shen, and Feng]{huang2022contrastive}
Zhicheng Huang, Xiaojie Jin, Chengze Lu, Qibin Hou, Ming-Ming Cheng, Dongmei Fu, Xiaohui Shen, and Jiashi Feng.
\newblock Contrastive masked autoencoders are stronger vision learners, 2022.

\bibitem[Bao et~al.(2022)Bao, Dong, Piao, and Wei]{bao2022beit}
Hangbo Bao, Li~Dong, Songhao Piao, and Furu Wei.
\newblock Beit: Bert pre-training of image transformers, 2022.

\bibitem[Mishra et~al.(2022)Mishra, Robinson, Chang, Jacobs, Sarna, Maschinot, and Krishnan]{mishra2022simple}
Shlok Mishra, Joshua Robinson, Huiwen Chang, David Jacobs, Aaron Sarna, Aaron Maschinot, and Dilip Krishnan.
\newblock A simple, efficient and scalable contrastive masked autoencoder for learning visual representations, 2022.

\bibitem[Mukhopadhyay et~al.(2023)Mukhopadhyay, Gwilliam, Yamaguchi, Agarwal, Padmanabhan, Swaminathan, Zhou, and Shrivastava]{mukhopadhyay2023textfree}
Soumik Mukhopadhyay, Matthew Gwilliam, Yosuke Yamaguchi, Vatsal Agarwal, Namitha Padmanabhan, Archana Swaminathan, Tianyi Zhou, and Abhinav Shrivastava.
\newblock Do text-free diffusion models learn discriminative visual representations?, 2023.

\bibitem[Hudson et~al.(2023)Hudson, Zoran, Malinowski, Lampinen, Jaegle, McClelland, Matthey, Hill, and Lerchner]{hudson2023soda}
Drew~A. Hudson, Daniel Zoran, Mateusz Malinowski, Andrew~K. Lampinen, Andrew Jaegle, James~L. McClelland, Loic Matthey, Felix Hill, and Alexander Lerchner.
\newblock Soda: Bottleneck diffusion models for representation learning, 2023.

\bibitem[Li et~al.(2022)Li, Chang, Mishra, Zhang, Katabi, and Krishnan]{li2022mage}
Tianhong Li, Huiwen Chang, Shlok~Kumar Mishra, Han Zhang, Dina Katabi, and Dilip Krishnan.
\newblock Mage: Masked generative encoder to unify representation learning and image synthesis, 2022.

\bibitem[Kalantidis et~al.(2020)Kalantidis, Sariyildiz, Pion, Weinzaepfel, and Larlus]{mochi}
Yannis Kalantidis, Mert~Bulent Sariyildiz, Noe Pion, Philippe Weinzaepfel, and Diane Larlus.
\newblock Hard negative mixing for contrastive learning.
\newblock \emph{Advances in neural information processing systems}, 33:\penalty0 21798--21809, 2020.

\bibitem[Robinson et~al.(2021)Robinson, Chuang, Sra, and Jegelka]{robinson2021contrastive}
Joshua~David Robinson, Ching-Yao Chuang, Suvrit Sra, and Stefanie Jegelka.
\newblock Contrastive learning with hard negative samples.
\newblock In \emph{International Conference on Learning Representations}, 2021.

\bibitem[Wang and Isola(2020)]{Wang2020UnderstandingCR}
Tongzhou Wang and Phillip Isola.
\newblock Understanding contrastive representation learning through alignment and uniformity on the hypersphere.
\newblock \emph{ArXiv}, abs/2005.10242, 2020.
\newblock URL \url{https://api.semanticscholar.org/CorpusID:218718310}.

\bibitem[Gwilliam and Shrivastava(2022)]{gwilliam2022supervised}
Matthew Gwilliam and Abhinav Shrivastava.
\newblock Beyond supervised vs. unsupervised: Representative benchmarking and analysis of image representation learning, 2022.

\bibitem[Adaloglou et~al.(2023)Adaloglou, Michels, Kalisch, and Kollmann]{TEMI_BMVC2023}
Nikolas Adaloglou, Felix Michels, Hamza Kalisch, and Markus Kollmann.
\newblock Exploring the limits of deep image clustering using pretrained models.
\newblock In \emph{34th British Machine Vision Conference 2023, {BMVC} 2023, Aberdeen, UK, November 20-24, 2023}, pages 297--299. {BMVA} Press, 2023.
\newblock URL \url{http://proceedings.bmvc2023.org/297/}.

\bibitem[Chang et~al.(2017{\natexlab{a}})Chang, Wang, Meng, Xiang, and Pan]{Chang_2017_ICCV}
Jianlong Chang, Lingfeng Wang, Gaofeng Meng, Shiming Xiang, and Chunhong Pan.
\newblock Deep adaptive image clustering.
\newblock In \emph{Proceedings of the IEEE International Conference on Computer Vision (ICCV)}, Oct 2017{\natexlab{a}}.

\bibitem[Huang et~al.(2023)Huang, Chen, Zhang, and Shan]{ProPos_IEEETrans2023}
Zhizhong Huang, Jie Chen, Junping Zhang, and Hongming Shan.
\newblock Learning representation for clustering via prototype scattering and positive sampling.
\newblock \emph{{IEEE} Trans. Pattern Anal. Mach. Intell.}, 45\penalty0 (6):\penalty0 7509--7524, 2023.
\newblock \doi{10.1109/TPAMI.2022.3216454}.
\newblock URL \url{https://doi.org/10.1109/TPAMI.2022.3216454}.

\bibitem[Dang et~al.(2021)Dang, Deng, Yang, Wei, and Huang]{Dang_2021_CVPR}
Zhiyuan Dang, Cheng Deng, Xu~Yang, Kun Wei, and Heng Huang.
\newblock Nearest neighbor matching for deep clustering.
\newblock In \emph{Proceedings of the IEEE/CVF Conference on Computer Vision and Pattern Recognition (CVPR)}, pages 13693--13702, June 2021.

\bibitem[Niu et~al.(2022)Niu, Shan, and Wang]{Niu_2022}
Chuang Niu, Hongming Shan, and Ge~Wang.
\newblock Spice: Semantic pseudo-labeling for image clustering.
\newblock \emph{IEEE Transactions on Image Processing}, 31:\penalty0 7264–7278, 2022.
\newblock ISSN 1941-0042.
\newblock \doi{10.1109/tip.2022.3221290}.
\newblock URL \url{http://dx.doi.org/10.1109/TIP.2022.3221290}.

\bibitem[Zhou and Zhang(2022)]{zhou2022deep}
Xingzhi Zhou and Nevin~L. Zhang.
\newblock Deep clustering with features from self-supervised pretraining, 2022.

\bibitem[He et~al.(2015)He, Zhang, Ren, and Sun]{he2015deep}
Kaiming He, Xiangyu Zhang, Shaoqing Ren, and Jian Sun.
\newblock Deep residual learning for image recognition, 2015.

\bibitem[Cao et~al.(2021)Cao, Brbic, and Leskovec]{cao2021open}
Kaidi Cao, Maria Brbic, and Jure Leskovec.
\newblock Open-world semi-supervised learning.
\newblock \emph{arXiv preprint arXiv:2102.03526}, 2021.

\bibitem[Wen et~al.(2023)Wen, Zhao, and Qi]{Wen_2023_ICCV}
Xin Wen, Bingchen Zhao, and Xiaojuan Qi.
\newblock Parametric classification for generalized category discovery: A baseline study.
\newblock In \emph{Proceedings of the IEEE/CVF International Conference on Computer Vision (ICCV)}, pages 16590--16600, October 2023.

\bibitem[Zhang et~al.(2023)Zhang, Khan, Shen, Naseer, Chen, and Khan]{Zhang_2023_CVPR}
Sheng Zhang, Salman Khan, Zhiqiang Shen, Muzammal Naseer, Guangyi Chen, and Fahad~Shahbaz Khan.
\newblock Promptcal: Contrastive affinity learning via auxiliary prompts for generalized novel category discovery.
\newblock In \emph{Proceedings of the IEEE/CVF Conference on Computer Vision and Pattern Recognition (CVPR)}, pages 3479--3488, June 2023.

\bibitem[Zhao et~al.(2023)Zhao, Wen, and Han]{GCP_ICCV2023}
Bingchen Zhao, Xin Wen, and Kai Han.
\newblock Learning semi-supervised gaussian mixture models for generalized category discovery.
\newblock In \emph{{IEEE/CVF} International Conference on Computer Vision, {ICCV} 2023, Paris, France, October 1-6, 2023}, pages 16577--16587. {IEEE}, 2023.
\newblock \doi{10.1109/ICCV51070.2023.01524}.
\newblock URL \url{https://doi.org/10.1109/ICCV51070.2023.01524}.

\bibitem[Ronen et~al.(2022)Ronen, Finder, and Freifeld]{deepDPM_CVPR2022}
Meitar Ronen, Shahaf~E Finder, and Oren Freifeld.
\newblock Deepdpm: Deep clustering with an unknown number of clusters.
\newblock In \emph{Proceedings of the IEEE/CVF Conference on Computer Vision and Pattern Recognition}, pages 9861--9870, 2022.

\bibitem[Chang et~al.(2017{\natexlab{b}})Chang, Wang, Meng, Xiang, and Pan]{dac_ICCV2017}
Jianlong Chang, Lingfeng Wang, Gaofeng Meng, Shiming Xiang, and Chunhong Pan.
\newblock Deep adaptive image clustering.
\newblock In \emph{2017 IEEE International Conference on Computer Vision (ICCV)}, pages 5880--5888, 2017{\natexlab{b}}.
\newblock \doi{10.1109/ICCV.2017.626}.

\bibitem[Fini et~al.(2021)Fini, Sangineto, Lathuili{\`e}re, Zhong, Nabi, and Ricci]{fini2021unified}
Enrico Fini, Enver Sangineto, St{\'e}phane Lathuili{\`e}re, Zhun Zhong, Moin Nabi, and Elisa Ricci.
\newblock A unified objective for novel class discovery.
\newblock In \emph{Proceedings of the IEEE/CVF International Conference on Computer Vision}, pages 9284--9292, 2021.

\bibitem[Wang et~al.(2024)Wang, Vaze, and Han]{sptnet_iclr2024}
Hongjun Wang, Sagar Vaze, and Kai Han.
\newblock Sptnet: An efficient alternative framework for generalized category discovery with spatial prompt tuning.
\newblock In \emph{International Conference on Learning Representations (ICLR)}, 2024.

\bibitem[Oquab et~al.(2023)Oquab, Darcet, Moutakanni, Vo, Szafraniec, Khalidov, Fernandez, Haziza, Massa, El-Nouby, et~al.]{dinov2}
Maxime Oquab, Timoth{\'e}e Darcet, Th{\'e}o Moutakanni, Huy Vo, Marc Szafraniec, Vasil Khalidov, Pierre Fernandez, Daniel Haziza, Francisco Massa, Alaaeldin El-Nouby, et~al.
\newblock Dinov2: Learning robust visual features without supervision.
\newblock \emph{arXiv preprint arXiv:2304.07193}, 2023.

\bibitem[Coates et~al.(2011)Coates, Ng, and Lee]{pmlr-v15-coates11a}
Adam Coates, Andrew Ng, and Honglak Lee.
\newblock An analysis of single-layer networks in unsupervised feature learning.
\newblock In Geoffrey Gordon, David Dunson, and Miroslav Dudík, editors, \emph{Proceedings of the Fourteenth International Conference on Artificial Intelligence and Statistics}, volume~15 of \emph{Proceedings of Machine Learning Research}, pages 215--223, Fort Lauderdale, FL, USA, 11--13 Apr 2011. PMLR.
\newblock URL \url{https://proceedings.mlr.press/v15/coates11a.html}.

\bibitem[Krizhevsky et~al.(2009)Krizhevsky, Hinton, et~al.]{krizhevsky2009learning}
Alex Krizhevsky, Geoffrey Hinton, et~al.
\newblock Learning multiple layers of features from tiny images.
\newblock 2009.

\end{thebibliography}
}

\FloatBarrier
\clearpage
\appendix

\section{Appendix / supplemental material}

\tableofcontents %
\addtocontents{toc}{\protect\setcounter{tocdepth}{2}}

\subsection{Preamble}
We give some details that we exclude from the main paper due to space constraints.
We give information about the datasets (Appendix~\ref{app:dataset_stats}) being used. Then we show a visualization of the hypothesis we use for the data distribution in the high dimensional space in Appendix~\ref{app:vis_of_2nd_nei_space}. Then we answer some frequently asked questions about the methodology being proposed with respect to novelty and fairness of comparison (Appendix~\ref{app:better_than_scan}, ~\ref{app:backbone_advantage}, ~\ref{app:dino_backbones}, ~\ref{app:novelty_nei}). We explain the empty spaces in clustering results in Appendix~\ref{app:empty_cells}. Then, we discuss the absence of contrastive loss in the SOTA model in Appendex~\ref{app:contrastive}. 
We describe hyperparameters for UNIC  in Appendix~\ref{app:heads}.
We also extend Figure~\ref{fig:sample-results} with more examples for GCD in Appendix~\ref{app:examples_gcd}.
Finally, we provide more examples for clustering in Appendix~\ref{app:examples_clustering}.

\subsection{Dataset statistics}
\label{app:dataset_stats}
\begin{table}[ht]
\caption{\textbf{Dataset Statistics} for our chosen tasks, Image Clustering and GCD.}
\label{tab:dataset-stats}
\resizebox{\linewidth}{!}{%
\begin{tabular}{@{}lcccccc@{}}
\toprule
                        & \multicolumn{3}{c}{Image Clustering} & \multicolumn{3}{c}{GCD}                  \\
Dataset      & \# Classes & Train Examples    & Test examples    & Old Classes & New Classes & Old labeled \% \\
\midrule
CIFAR-10~\cite{krizhevsky2009learning}       &10           & -                  &-                  &   10          &     10        &       50\%       \\
STL-10~\cite{pmlr-v15-coates11a}       &10            & 5k                  &8k                  &   -          &     -        &       -       \\
ImageNet-50~\cite{5206848}  &   50         &    64k               &  2.5k                &        -     &   -          &     -         \\
ImageNet-100~\cite{5206848} &  100          &        128k           &      5k            &     50        &     50        &   50\%           \\
ImageNet-200~\cite{5206848} &   200         &    256k               &       10k           &         -    &         -    &       -    \\
CUB-200 & 200 & - & - & 100 & 100 & 50\% \\
Aircrafts & 200 & - & - &  98 &  98 & 50\% \\
SCars & 100 & - & - & 50 & 50 & 50\% \\
\bottomrule
\end{tabular}%
}
\end{table}

Table~\ref{tab:dataset-stats} contains the information about all the datasets being used in this work. It also shows how Old/New splits are made for GCD.

\subsection{Visualization of Second Order Neighborhoods}
\label{app:vis_of_2nd_nei_space}

The mechanism for cleaning neighbors based on second-order neighborhoods has been introduced in Figure~\ref{fig:mining-diagram} and Equation~\ref{eq:clearning}. In addition, this intuitive idea is experimentally shown in Figure~\ref{fig:union_removal} with real data statistics. In this subsection, we try to visualize the phenomena within the Imagenet-100 dataset. The objective of this analysis is to remove the false negatives within first-order neighbors. The data points which are in the decision boundary of clusters will have this issue. However, these data points on the decision boundary should be identified without having access to the ground truth labels.

In order to demonstrate this, we pick two classes, embedd the data points onto the DINO space (which is 768 dimensional), and perform dimension reduction up to 2 dimensions (using PCA) for visualization. We try to show important information here.
\begin{itemize}
    \item The first-order neighborhood of datapoints inside clusters will have high-quality (less noisy) positive neighbors.
    \item The second-order neighborhood size is a good proxy to figure out the data points at the decision boundaries.
\end{itemize}

\begin{figure*}[ht]
    \centering
    \includegraphics[width=\textwidth]{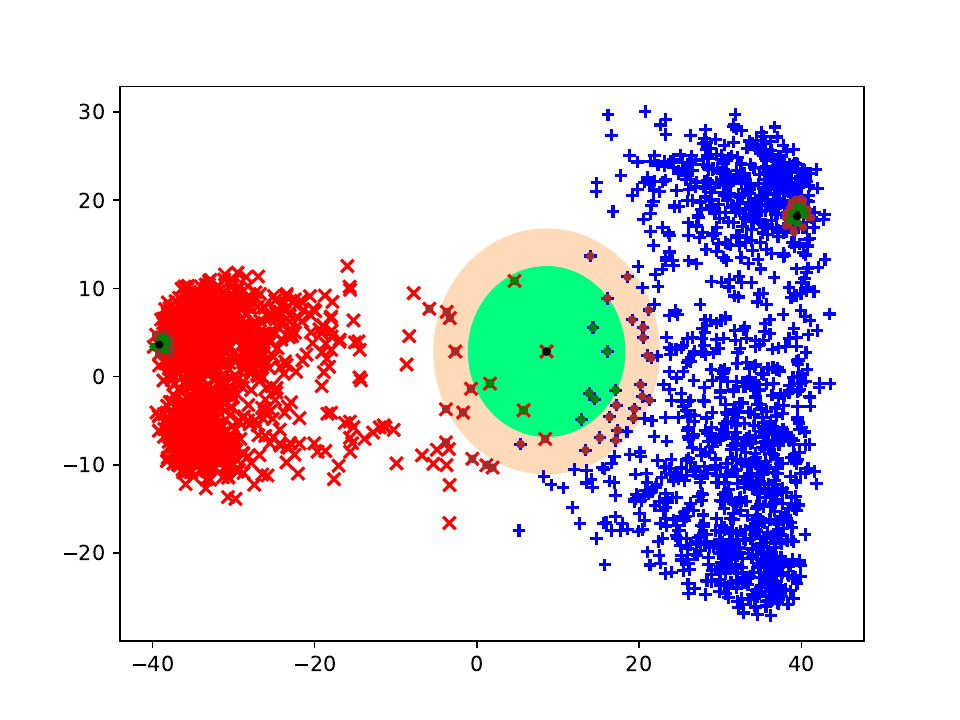}
    \caption{Neighbor cleaning based on second order information}
    \label{fig:union_neighbor_purity2}
\end{figure*}

Fig \ref{fig:union_neighbor_purity2} shows the embedding space for two classes in Imagenet 100 dataset. The class labels are denoted by blue pluses \begin{tikzpicture}[baseline=-0.8ex] \draw[blue, line width=0.4ex] (-0.8ex,0) -- (0.8ex,0);    \draw[blue, line width=0.4ex] (0,-0.8ex) -- (0,0.8ex);  \end{tikzpicture} and red crosses \begin{tikzpicture}[baseline=-0.8ex]     \draw[red, line width=0.3ex] (-0.7ex,-0.7ex) -- (0.7ex,0.7ex);    \draw[red, line width=0.3ex] (0.7ex,-0.7ex) -- (-0.7ex,0.7ex);  \end{tikzpicture} for the reader. Please note that they are \textbf{not} available for the algorithm. Consider three image embeddings denoted in black circles (left \begin{tikzpicture}[baseline=-0.8ex]     \draw[red, line width=0.3ex] (-0.7ex,-0.7ex) -- (0.7ex,0.7ex);    \draw[red, line width=0.3ex] (0.7ex,-0.7ex) -- (-0.7ex,0.7ex);      \fill[black] (0,0) circle (0.5ex); \end{tikzpicture}, middle \begin{tikzpicture}[baseline=-0.8ex]
    \draw[red, line width=0.3ex] (-0.7ex,-0.7ex) -- (0.7ex,0.7ex);
    \draw[red, line width=0.3ex] (0.7ex,-0.7ex) -- (-0.7ex,0.7ex);
    \fill[black] (0,0) circle (0.5ex);
\end{tikzpicture} and right \begin{tikzpicture}[baseline=-0.8ex] \draw[blue, line width=0.4ex] (-0.8ex,0) -- (0.8ex,0);    \draw[blue, line width=0.4ex] (0,-0.8ex) -- (0,0.8ex);    \fill[black] (0,0) circle (0.5ex);
\end{tikzpicture}). The first order neighbors are denoted in green circles \begin{tikzpicture}\definecolor{darkgreen}{rgb}{0.0, 0.5, 0.0} \fill[darkgreen] (0,0) circle (0.5ex); \end{tikzpicture} and second order neighbors are denoted in brown circles \begin{tikzpicture}\definecolor{brown}{rgb}{0.65, 0.16, 0.16} \fill[brown] (0,0) circle (0.5ex); \end{tikzpicture}. The second-order union sizes of left and right are small enough for them to be included. Therefore, the first order neighborhoods of left and right are included. The second-order union size of the middle \begin{tikzpicture}[baseline=-0.3ex]
    \definecolor{lightgreen}{rgb}{0.56, 0.93, 0.56}
    \definecolor{lightbrown}{rgb}{0.87, 0.72, 0.53}
    
    \fill[lightgreen] (0,0) rectangle (1.5ex,1.5ex);
    
    \fill[lightbrown] (1.5ex,0) rectangle (3ex,1.5ex);
\end{tikzpicture} is large so that the first order neighborhood of the middle \begin{tikzpicture}[baseline=-0.3ex]
    \definecolor{lightgreen}{rgb}{0.56, 0.93, 0.56}
    \definecolor{lightbrown}{rgb}{0.87, 0.72, 0.53}
    
    \fill[lightgreen] (0,0) rectangle (1.5ex,1.5ex);
\end{tikzpicture} is excluded. This cleaning startergy is effective because the green region of left contains red crosses only \begin{tikzpicture}[baseline=-0.8ex]\definecolor{darkgreen}{rgb}{0.0, 0.5, 0.0}     \draw[red, line width=0.3ex] (-0.7ex,-0.7ex) -- (0.7ex,0.7ex);    \draw[red, line width=0.3ex] (0.7ex,-0.7ex) -- (-0.7ex,0.7ex);      \fill[darkgreen] (0,0) circle (0.5ex); \end{tikzpicture} and the green region of right contains blue pluses \begin{tikzpicture}[baseline=-0.8ex] \definecolor{darkgreen}{rgb}{0.0, 0.5, 0.0} \draw[blue, line width=0.4ex] (-0.8ex,0) -- (0.8ex,0);    \draw[blue, line width=0.4ex] (0,-0.8ex) -- (0,0.8ex);    \fill[darkgreen] (0,0) circle (0.5ex);
\end{tikzpicture} only. The green region of middle contains both croses and pluses \begin{tikzpicture}[baseline=-0.8ex]
    \definecolor{darkgreen}{rgb}{0.0, 0.5, 0.0}  
    \definecolor{lightgreen}{rgb}{0.56, 0.93, 0.56}    
    \fill[lightgreen] (-1ex,-1ex) rectangle (3.5ex,1.5ex);
    \draw[red, line width=0.3ex] (-0.7ex,-0.7ex) -- (0.7ex,0.7ex);    
    \draw[red, line width=0.3ex] (0.7ex,-0.7ex) -- (-0.7ex,0.7ex);      
    \fill[darkgreen] (0,0) circle (0.5ex);  
    \draw[blue, line width=0.4ex] (1.2ex,0) -- (2.8ex,0);    
    \draw[blue, line width=0.4ex] (2ex,-1ex) -- (2ex,0.8ex);    
    \fill[darkgreen] (2ex,0) circle (0.5ex); 
    
\end{tikzpicture} . Therefore, excluding this reduces noise in positive neighbors.

\subsection{FAQ(1) Is this better than SCAN+DINO?}
\label{app:better_than_scan}

SCAN~\citep{SCAN_ECCV2020} is a seminal work in Deep Clustering. We acknowledge that certain components of our work ($L_{POS}$, $L_{ENT}$) are inspired by SCAN.

SCAN was published in 2020 with a ResNet backbone. The absence of numbers in ImageNet columns for most ResNet based work in Table~\ref{tab:results_clustering} shows that SCAN was one of the few work that was scaleable enough for larger datasets beyond 100 classes. 

Few recent methods \citep{TEMI_BMVC2023,zhou2022deep} have been proposed for image clustering using ViT-B/16 backbone. They have outperformed earlier ResNet work by a large margin.

Firstly, we show running kMeans on ViT-B/16 features is enough to outperform every work in the literature for Imagenet (subset) scale datasets. Then, we implement SCAN with ViT-B/16 and show it would be the state of the art in most clustering tasks even today.

However, the last row of Table~\ref{tab:results_clustering} clearly shows that our work achieves better performance than SCAN+ViT-B/16.

In addition, we expand on the differences between SCAN and our work in Table~\ref{tab:scan_to_unic}. UNIC has a sequence of design decisions that improves the performance with respect to SCAN. Also, the utility of multiple heads in SCAN is not scalable when it comes to tuning the ViT backbone due to higher memory requirements.

\begin{table}[h]

\caption{\textbf{UNIC Ablation on ImageNet-50}}
\label{tab:scan_to_unic}
\begin{center}
    
\resizebox{0.5\textwidth}{!}{
\begin{tabular}{@{}ll ccc@{}}
\toprule
\multicolumn{2}{c}{Algorithm}  & ACC & NMI & ARI  \\ \midrule
$L_{\text{SCAN}}$ & +Resnet + heads & \multirow{2}{*}{76.8} & \multirow{2}{*}{82.2} & \multirow{2}{*}{66.1} \\
\multicolumn{2}{c}{ = SCAN \citep{SCAN_ECCV2020}}\\
\midrule
$L_{\text{SCAN}}$  & +VitB/16 + heads  & 85.48 & 88.58 & 78.19 \\
$L_{\text{SCAN}}$  & +VitB/16 & 83.33 & 88.65 & 76.81  \\
$L_{\text{SCAN}}$  & +VitB/16 + heads + Tune & \multicolumn{3}{c}{Memory error} \\
$L_{\text{SCAN}}$  & +VitB/16 + Tune & 87.12 & 90.18 & 80.45 \\
$L_{\text{SCAN}}$  & +VitB/16 + Tune + $L_{\text{neg}}$ & 89.04 & 90.70 & 82.19 \\
\midrule
$L_{\text{SCAN}}$ & +VitB/16 + Tune + cleaning & \multirow{2}{*}{90.80} & \multirow{2}{*}{91.81} & \multirow{2}{*}{84.25} \\
\multicolumn{2}{c}{ = UNIC}\\
\bottomrule

\end{tabular}
}
\end{center}

\end{table}

\subsection{FAQ(2) What is the novelty of this paper if Neighbor Information has already been used for Clustering?}
\label{app:novelty_nei}

Almost every work in clustering literature has used positive neighbors mined by the Euclidean distance in embedding space. However, our work improves upon this idea on multiple fronts. Positive neighbor mining is cleaned by thresholding with respect to second-order neighborhood sizes. It should be noted that previous work \citep{SCAN_ECCV2020,TEMI_BMVC2023} has run simulations with perfect positive neighbors (assuming they have an oracle to get this information) and showed that it would improve the clustering performance. However, UNIC is the first attempt at getting there. In addition, negative neighbor mining is also a novel contribution in utilizing neighbor informaiton.

\subsection{FAQ(3) Is it possible to attribute the performance gains to new backbones and pertaining only?}
\label{app:backbone_advantage}

The naive performance of the backbone for both clustering and GCD can be seen with the kMeans result. The gain over the kMeans result is what a particular work contributes to. Interestingly, when the backbones are of moderate performance, there is much room for a proposed algorithm to contribute. This is seen by the larger gains among the ResNet based algorithms in Table~\ref{tab:results_clustering}. However, when the backbone has great performance, algorithms should be designed carefully to attain performance gains. Our results in Table~\ref{tab:results_clustering} and Table~\ref{tab:gcd_results} shows how UNIC has performance gains over kMeans in this challenging situation as well.

\subsection{FAQ(4) Why are there many empty cells in the clustering results table?}
\label{app:empty_cells}

Earlier deep clustering methods were not scale-able for a large number of classes. They usually attempted to solve CIFAR-10 like datasets or subsets of CIFAR-100 (20 classes out of the full 100), and Imagenet (10 classes out of the full 1000). Scaling them to present day benchmarks (up to 200 classes) requires significant engineering effort.

\subsection{Fair comparison with respect to backbones}
\label{app:dino_backbones}
Unsupervised and semi-supervised (including GCD) work depends on a pre-trained backbone to bootstrap their frameworks. It is impossible to start with a randomly initialized backbone. The performance of an algorithm is highly dependent on the backbones being used. 

Firstly, the fairness of comparison aspect of choosing backbones will be considered. Image clustering is an older problem that was first explored with ResNet backbones since 2020. The performance of the models made a significant jump once ViT backbones started being used since 2022 for the task. Inorder to provide a fair comparison, we re-implement important work from ResNet years (SCAN~\citep{SCAN_ECCV2020}) with a ViT-B/16 backbone pretrained with DINOv1. There are no clustering papers that uses a DINOv2 backbone.

GCD was defined more recently in \citet{GCD_CVPR2022}. Therefore, there is no ResNet-based work on the problem. Earlier work reported SOTA with ViT-B/16 DINOv1. More recent papers started reporting numbers for ViT-B/16 DINOv2 (with clear gains over DINOv1 numbers). We report our numbers with both DINOv1 and DINOv2 backbones. Our method outperforms previous methods with +1.45\% accuracy on Imagenet-100 with DINOv1. Our method achieves SOTA over all (older and newer) methods with DINOv2 backbone at +0.75\% on Imagenet-100 and +5.06\% on CUB-200. It should be noted that DINOv2 CIFAR-10 numbers were not reported in the literature and we have re-run their code to generate those numbers.

Secondly, the backbone also dictates which type of frameworks (or components of frameworks) can be applied. For example, multiple heads and self-labelling like ideas from SCAN cannot be applied to ViT-B/16 DINO backbones. We have recreated SCAN for modern backbones by carefully changing these components of the original algorithm. Similarly, modern work like TEMI~\citep{TEMI_BMVC2023} reports very low numbers for older ResNet MoCo backbones.

\subsection{Contrastive loss}
\label{app:contrastive}

Earlier work in the domain has used Supervised Contrastive loss \citep{GCD_CVPR2022} to improve the performance. We conduct experiments with a similar setting and show it is redundant for the tasks being studied in this work. The results are given in Table~\ref{tab:ablation_commponents}.

For this ablation, we use the Supervised Contrastive loss coupled with the Dino-head projection of backbone features. We use the weighting term 5.0 to add this to the overall loss function and jointly train the full pipeline.

\subsection{Hyperparameters}
\label{app:heads}

\textbf{Neighbors: }
We show experiments for different values of $\tau_1$, $\tau_2$ and $\eta$.  We fix $\tau_2$ as 6300 for ImageNet splits, 1000 for STL and CUB datasets, and 10,000 for CIFAR10 based on heuristics. We fix $\tau_1=10$, $\eta=70$ for all the experiments except for ImageNet splits in Clustering setting. For that, we use $\tau_1=50$, $\eta=1500$

\textbf{Batch size: }
Figure~\ref{fig:hyperparameter_convergence} analyzes ImageNet-50 clustering performance of the proposed system with respect to three hyperparameters -- batch sizes, clustering head architectures, and freezing strategies. This subsection describes the experiments in detail.

Our base experiment runs on a batch size of 128 to be consistent with the rest of the GCD literature. 
The ablations results for batch sizes 256 and 512 are inferior to 128. 
However, it should be noted that the ImageNet-200 clustering result was obtained for the batch size of 512.
Batches of smaller sizes are not representative of the full dataset statistics when it comes to a 200-class case. 
Specifically, this is detrimental to the optimization of the entropy-based loss function.

\textbf{Clustering heads: }
We experiment with three clustering heads. 
All these take Dino ViT-B/16 features (unnormalized) as input. 
The base experiments are run with the 2-layer perceptron (abbreviated MLP) with an intermediate dimension 2048. 
FC refers to the single, fully-connected linear layer. 
SA refers to a self-attention-based classification architecture with patch size $16 \times 16$ and intermediate MLP dimension 2048.

\textbf{Trainable parameters: }
In terms of training the backbone itself, we find it helps to partially finetune it (see Figure~\ref{fig:hyperparameter_convergence}).
We find that freezing the whole backbone (training none of it) is too restrictive.
We find that fully-finetuning is also suboptimal.
So instead, we train only the final ($\text{12}^\text{th}$) transformer block.

\subsection{FAQ: Why are mined negatives better than labeled negatives in Table~\ref{tab:ablation_gcd}?}
\label{app:labelled-negatives}

Labeled negatives will always have a better true negative percentage than mined negatives. Therefore, it is counter-intuitive to witness a result where the All Class Accuracy using the mined negatives (83.22\%) is better than the All Class Accuracy using the labeled negatives (82.66\%).

Firstly, it should be noted that the Old Class Accuracy of 92.28\% in the labeled case is above 91.97\% in the mined case. This result matches the primary intuition about the lower number of false negatives in the labeled case.

The counterintuitive drop comes when the New Class Accuracy drops from 78.82\% of the mined case to 77.83\% in the labeled case. Consider a dataset $D$ with $N$ images. $0.25N$ will be labelled old, $0.25N$ will be unlabelled old, and $0.5N$ will be unlabelled new. Consider the case where every image gets $k$ negative neighbors. 

Given the label agnostic mining, the unlabelled $0.75N$ will get an equal proportion of negative neighbors from Old and New classes. This will amount to $0.375Nk$ from each. Consider that $\alpha$ fraction of negative neighbors for the labeled images come from the rest of the labeled images (all true negatives). However, this set will only have images from Old classes. Therefore, $(1 - \alpha)$ fraction should be mined. This will have equal fractions of Old and New classes. However, in total, $0.25Nk\left(\alpha + \frac{1 - \alpha}{2}\right)$ of the negative neighbors will be from Old classes and  $0.25Nk\left(\frac{1 - \alpha}{2}\right)$ will be from New classes. For any choice of $\alpha$ (our reported result is for 0.5), the Old classes will be overrepresented as negative neighbors in the training. The under representation of the New class is the major contributing factor to the reported accuracy drop for New Classes, and therefore for the overall accuracy.

\subsection{FAQ: How Does UNIC's Neighbor Mining Compare to Hard Negative Mining}

MoCHi \cite{mochi} and HCL \cite{robinson2021contrastive} has demonstrated the effectiveness of an unsupervised hard negative mining step on top of the vanilla SSL pretraining. These works shows impressive image classification results for the fully supervised case compared to the vanilla SSL pretraining. This appendix subsection attempts to explore (a) whether such techniques are useful for UNIC's problem settings, and (b) the reasons for it.

Both HCL (built on SimCLR~\cite{chen2020simple}) and MoCHi (built on MoCoV2~\cite{chen2020improved}) attempt to find hard negatives to perform contrastive learning. Their objective is for the images in the dataset to uniformly be spread out through the embedding space. To this end, MoCHI specifically measures “uniformity” as defined in \cite{Wang2020UnderstandingCR}. HCL calls this “optimal embedding” and proves theoretical guarantees.

Such uniform spread of embeddings can be classified using MLPs trained with fully supervised learning. This is why HCL and MoCHi outperform the SimCLR and MoCo on classification tasks.
It should be noted that spreading out the dataset uniformly is detrimental to the naturally occurring decision boundaries (i.e. cluster structure) that are used in unsupervised and semi-supervised learning settings. Our design choices in UNIC are to do the opposite – bring embeddings of the same class together while increasing the distance between the embeddings for different classes. In essence, this creates clusters (with natural decision boundaries) instead of a uniform spread.

We experiment with unsupervised kMeans clustering on HCL and MoCHi embeddings and append the results in Table 1. It should be noted that these numbers are below the kMeans result from SimCLR for STL-10 (which is reported in the first row from \cite{SCAN_ECCV2020}) and kMeans result from MoCoV2 for all datasets. Then, we run the full UNIC training pipeline on the MoCHi and MoCoV2 backbones and report results in Table~\ref{tab:different_backbones}. MoCHi performs worse than vanilla MoCoV2 as the base representation for UNIC. 

All these experiments provide strong evidence of why contrastive learning with hard negatives is less desirable for UNIC use cases.

\subsection{UNIC with other backbones}

The main text of the paper experiments with DINOv1 and DINOv2 backbones for UNIC and compare with the current state of the art. This allows us to evaluate the proposed solution on it's merits rather than getting an unfair advantage from the backbone's pretraining. The fairness aspect of this is explained in Appendix~\ref{app:dino_backbones}. In this subsection, we run the UNIC pipeline with different backbones and report results in Table~\ref{tab:different_backbones}.

The results from Table~\ref{tab:different_backbones} shows how UNIC can be used on top of every SSL-trained backbone we have experimented on. UNIC reports consistent gain over the kMeans result from the vanilla backbone for most cases. However, the gains are minimal when the vanilla backbone's performance is $< 65\%$. UNIC has significant gains for all better backbones. It should be noted that ViT backbones are generally better than ResNet backbones. Also, the best performance of UNIC is obtained when the backbone is supervised-trained.

These experimental results are strong evidence for UNIC's generalizability over a spectrum of weak backbones to very strong backbones.

\begin{table}[h]
\begin{center}

\caption{\textbf{UNIC Clustering Performance with Different Backbones on ImageNet-50}\label{tab:different_backbones}}

\resizebox{0.45\textwidth}{!}{%

\begin{tabular}{@{}ll ccc ccc@{}}
\toprule
& &\multicolumn{3}{c}{kMeans} & \multicolumn{3}{c}{UNIC} \\ 
\cmidrule(l){3-5}
\cmidrule(l){6-8}
Pretraining  & Backbone & ACC & NMI & ARI & ACC & NMI & ARI \\ \midrule
MOCHi  ~\cite{mochi} & ResNet-50 & 61.88 & 73.44 & 44.92 & 61.80 & 72.81 & 48.80  \\
MoCOV2  ~\cite{chen2020improved} & ResNet-50 & 63.04 & 75.75 & 47.00 & 64.28 & 76.64 & 53.42  \\
SwAV  ~\cite{caron2020unsupervised} & ResNet-50 & 65.32 & 74.27 & 47.07 & 73.20 & 80.74 & 61.86  \\
DINOv1  ~\cite{caron2021emerging} & ResNet-50 & 70.68 & 77.82 & 52.17 & 75.44 & 81.62 & 64.77  \\
DINOv1  ~\cite{caron2021emerging} & VitB/16 & 82.36 & 87.91 & 73.89 & 90.80 & 91.81 & 84.25  \\
iBot  ~\cite{zhou2022ibot} & VitB/16 & 82.48 & 86.94 & 68.10 & 85.04 & 89.45 & 77.70  \\
DINOv2  ~\cite{dinov2} & VitB/14 & 87.20 & 89.22 & 60.42 & 94.60 & 95.09 & 90.77  \\
Supervised  ~\cite{dosovitskiy2021image} & VitB/16 & 89.76 & 90.12 & 66.08 & 96.44 & 95.99 & 93.18  \\
\bottomrule
\end{tabular}%
}
\end{center}
\end{table}

\subsection{Visual Examples GCD}
\label{app:examples_gcd}

\begin{figure*}[]
    \centering
    \includegraphics[width=\textwidth]{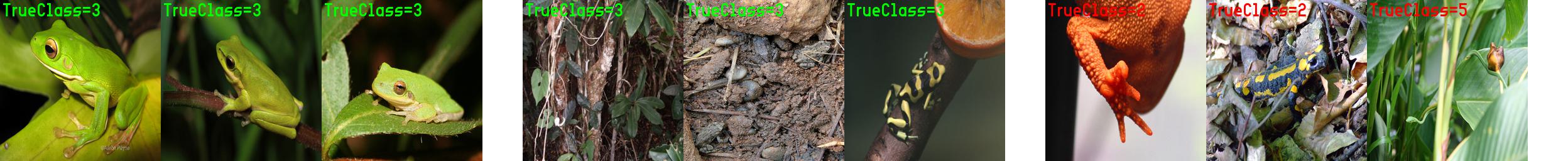}
    \includegraphics[width=\textwidth]{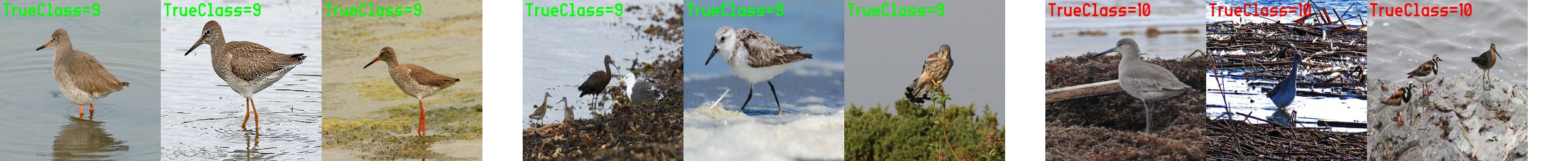}
    \includegraphics[width=\textwidth]{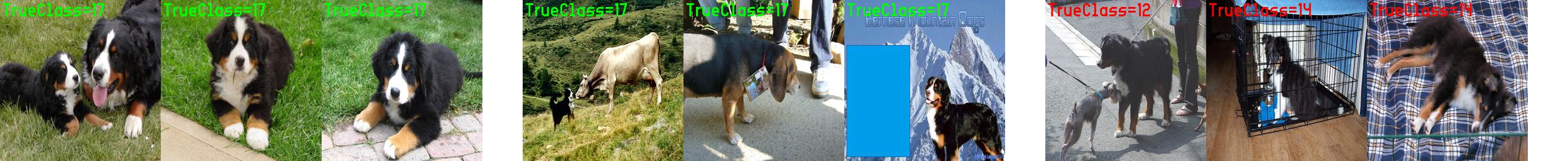}
    \includegraphics[width=\textwidth]{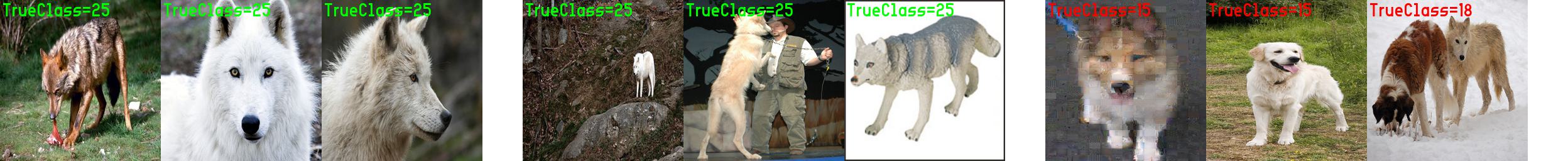}
    \includegraphics[width=\textwidth]{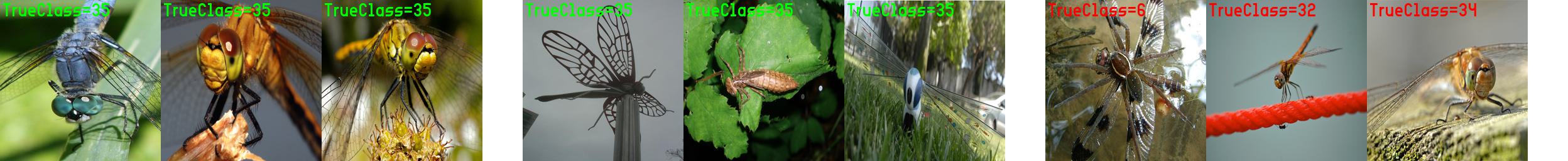}
    \caption{Example images from ImageNet-100 GCD for old classes. Left : highest confident true positives; Middle: least confident true positives; Right: False positives.}
    \label{fig:sup-images-gcd-old}
\end{figure*}

\begin{figure*}[]
    \centering
    \includegraphics[width=\textwidth]{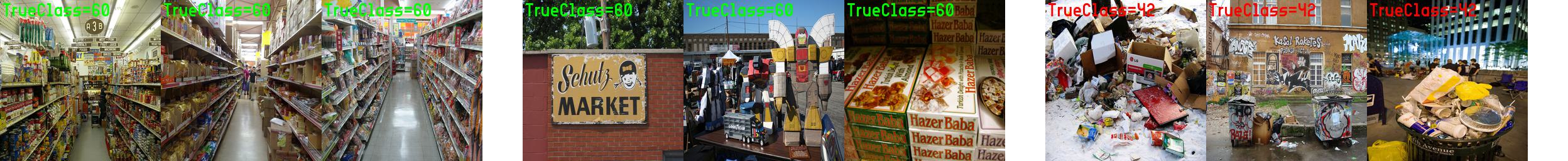}
    \includegraphics[width=\textwidth]{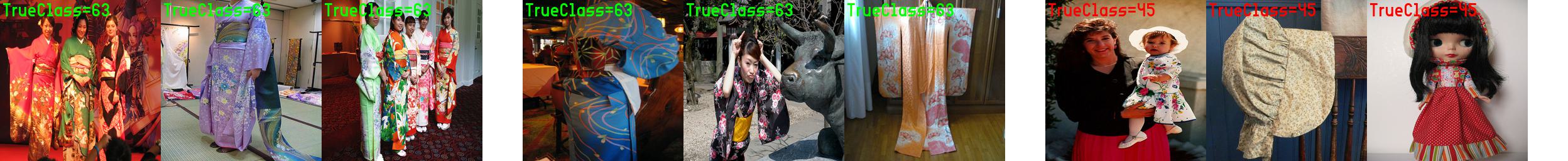}
    \includegraphics[width=\textwidth]{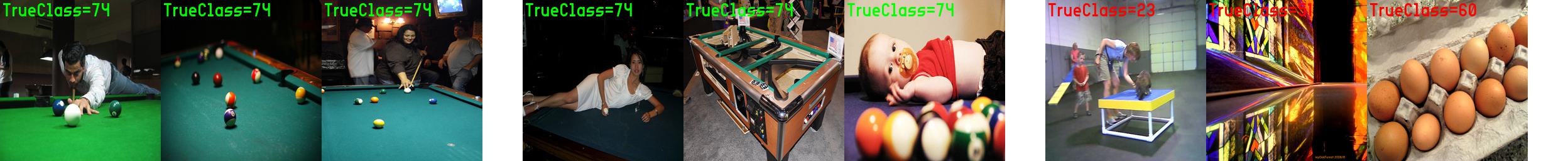}
    \includegraphics[width=\textwidth]{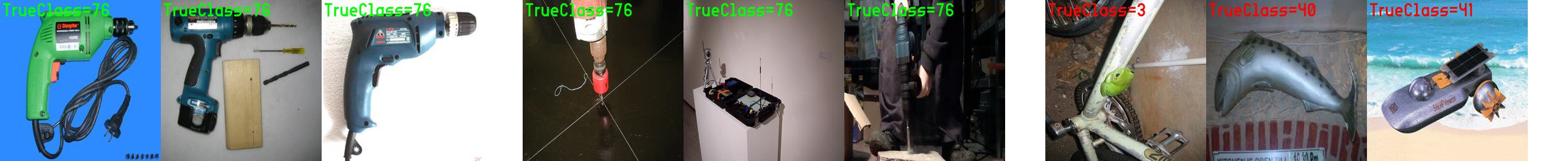}
    \includegraphics[width=\textwidth]{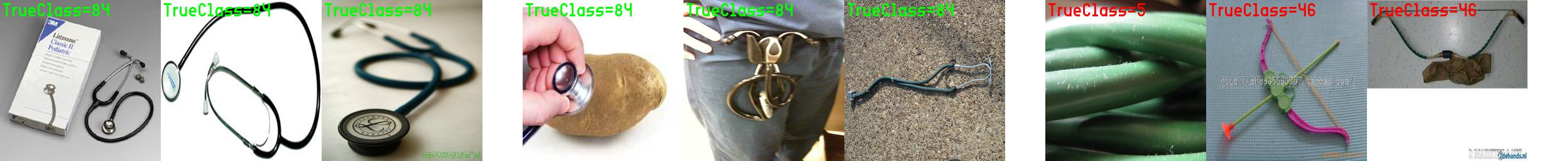}
    \caption{Example images from ImageNet-100 GCD for new classes. Left : highest confident true positives; Middle: least confident true positives; Right: False positives.}
    \label{fig:sup-images-gcd-new}
\end{figure*}

Some visual examples for GCD task on the ImageNet-100 dataset is given in this section. 
Figure~\ref{fig:sup-images-gcd-old} gives examples for old classes from GCD. 
Similarly, Figure~\ref{fig:sup-images-gcd-new} has examples from new classes.

\subsection{Visual Examples -- Clustering}
\label{app:examples_clustering}
\begin{figure*}[t]
    \centering
    \includegraphics[width=\textwidth]{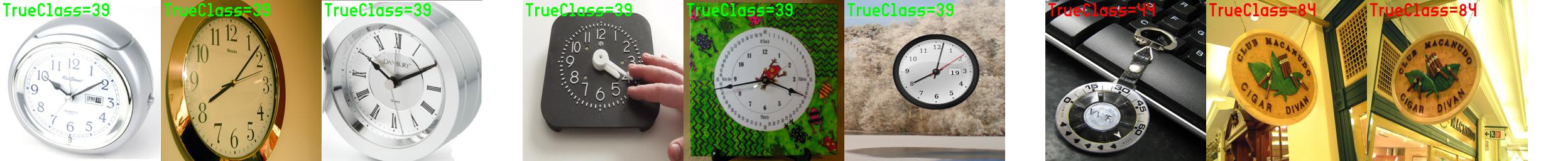}
    \includegraphics[width=\textwidth]{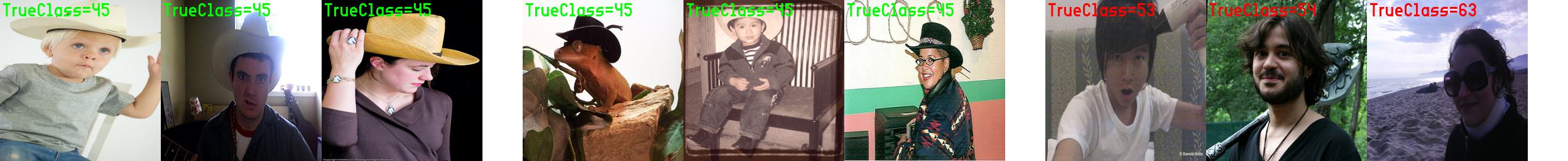}
    \includegraphics[width=\textwidth]{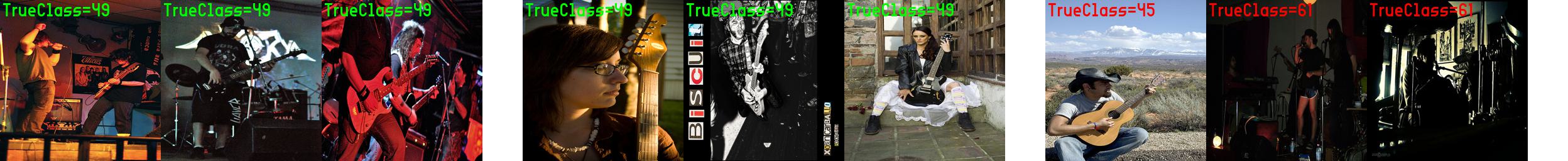}
    \includegraphics[width=\textwidth]{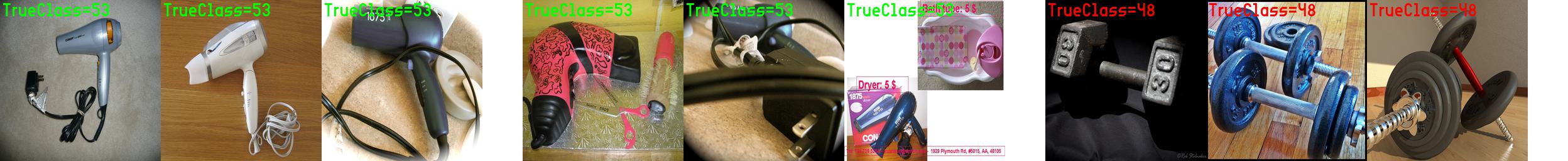}
    \includegraphics[width=\textwidth]{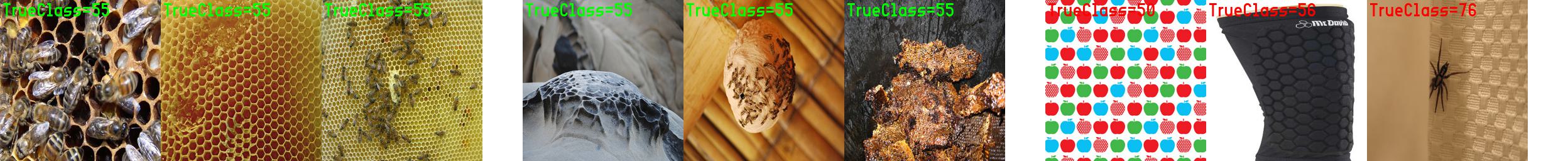}
    \includegraphics[width=\textwidth]{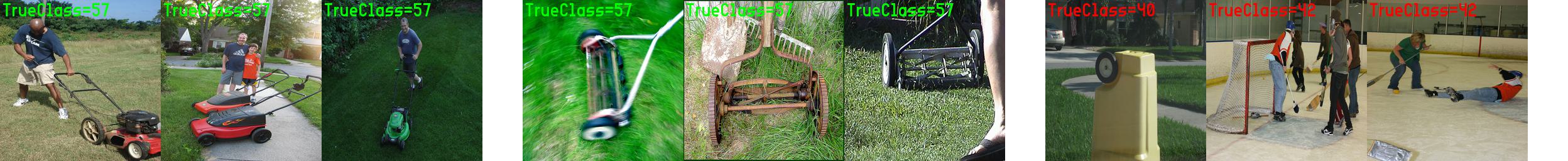}
    \includegraphics[width=\textwidth]{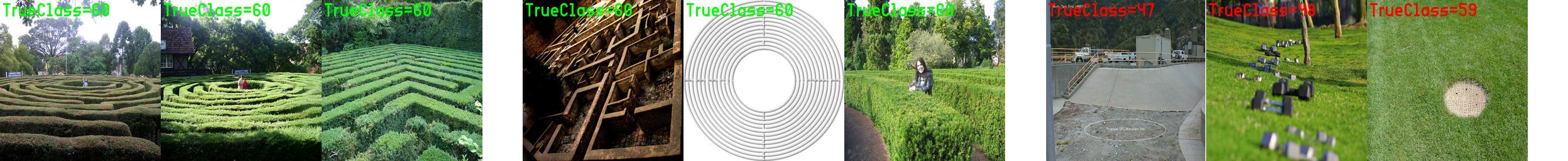}
    \includegraphics[width=\textwidth]{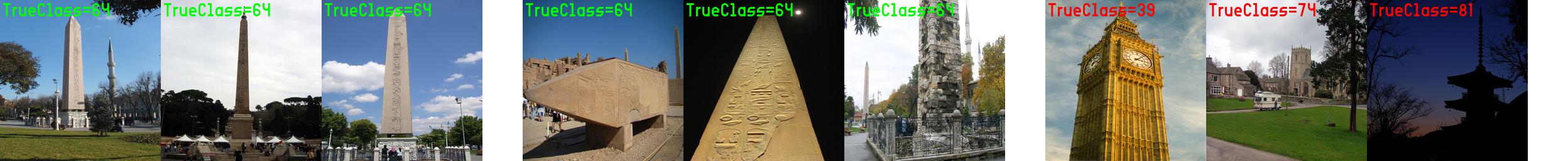}
    \raggedright \includegraphics[width=0.795\textwidth]{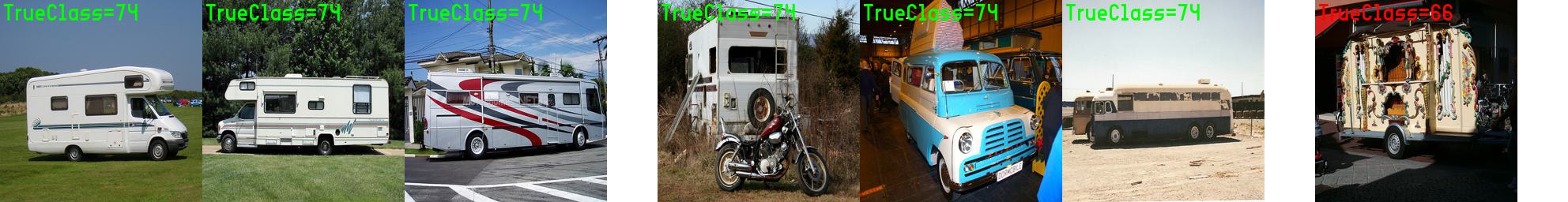} 
    \includegraphics[width=\textwidth]{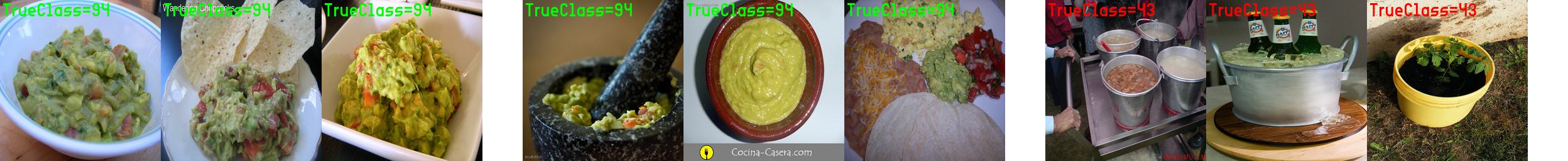}
    
    \caption{Example images from ImageNet-100 clustering. Left : highest confident true positives; Middle: least confident true positives; Right: False positives. \label{fig:sup-images-clustering}}
    
\end{figure*}

Figure~\ref{fig:sup-images-clustering} gives more examples for ImageNet-100 clusters learned in the fully-unsupervised setting.

\end{document}